\documentclass[preprint,12pt]{elsarticle}

\usepackage{amssymb}
\usepackage{amsthm}
\usepackage{graphicx}
\usepackage{subfigure}

\usepackage{amsmath, amsfonts}
\usepackage{algorithm}
\usepackage{algorithmic}
\usepackage{enumerate}
\usepackage{flushend}
\usepackage{booktabs}
\usepackage{longtable}
\usepackage{setspace}
\usepackage{verbatim}
\usepackage{url}
\usepackage{color}
\newcommand{\ws}[1]{\textcolor{blue}{#1}}
\usepackage{ulem}
\usepackage{makecell}
\usepackage{multirow}

\journal{Neural Networks}

\begin{document}

\begin{frontmatter}



\title{Self-grouping Convolutional Neural Networks}

\author[mymainaddress,mysecondaryaddress]{Qingbei~Guo}

\author[mysecondaryaddress]{Xiao-Jun~Wu\corref{mycorrespondingauthor}}
\cortext[mycorrespondingauthor]{Corresponding author}
\ead{wu\_xiaojun@jiangnan.edu.cn}

\author[mythirdaddress]{Josef~Kittler\fnref{myIEEEauthor}}
\fntext[myIEEEauthor]{life Member,~IEEE}

\author[mysecondaryaddress]{Zhiquan~Feng}

\address[mymainaddress]{Jiangsu Provincial Engineering Laboratory of Pattern Recognition and Computational Intelligence, Jiangnan University, Wuxi 214122, China}
\address[mysecondaryaddress]{Shandong Provincial Key Laboratory of Network based Intelligent Computing, University of Jinan, Jinan 250022, China}
\address[mythirdaddress]{Centre for Vision, Speech and Signal Processing, University of Surrey, Guildford GU2 7XH, UK}

\begin{abstract}
Although group convolution operators are increasingly used in deep convolutional neural networks to improve the computational efficiency and to reduce the number of parameters, most existing methods construct their group convolution architectures by a predefined partitioning of the filters of each convolutional layer into multiple regular filter groups with an equal spatial group size and data-independence, which prevents a full exploitation of their potential. To tackle this issue, we propose a novel method of designing self-grouping convolutional neural networks, called SG-CNN, in which the filters of each convolutional layer group themselves based on the similarity of their importance vectors. Concretely, for each filter, we first evaluate the importance value of their input channels to identify the importance vectors, and then group these vectors by clustering. Using the resulting \emph{data-dependent} centroids, we prune the less important connections, which implicitly minimizes the accuracy loss of the pruning, thus yielding a set of \emph{diverse} group convolution filters. Subsequently, we develop two fine-tuning schemes, i.e. (1) both local and global fine-tuning and (2) global only fine-tuning, which experimentally deliver comparable results, to recover the recognition capacity of the pruned network. Comprehensive experiments carried out on the CIFAR-10/100 and ImageNet datasets demonstrate that our self-grouping convolution method adapts to various state-of-the-art CNN architectures, such as ResNet and DenseNet, and delivers superior performance in terms of compression ratio, speedup and recognition accuracy. We demonstrate the ability of SG-CNN to generalise by transfer learning, including domain adaption and object detection, showing competitive results. \ws{Our source code is available at \url{https://github.com/QingbeiGuo/SG-CNN.git}.}
\end{abstract}


\end{frontmatter}


\section{Introduction}\label{sec:Introduction}

\begin{figure*}
\begin{minipage}{1\textwidth}
  \centering
  \subfigure[]{
  \includegraphics[trim=5mm 0mm -5mm 0mm, width=0.93in]{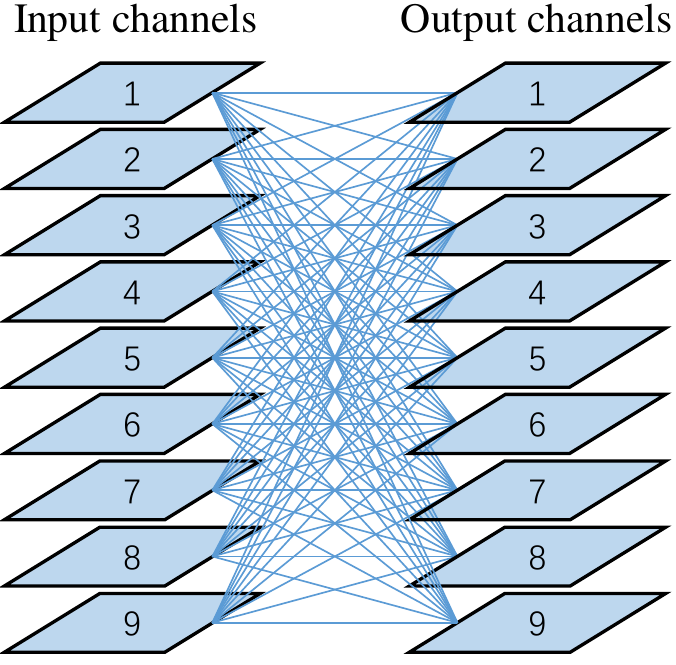}}
  \hspace{0in}
  \subfigure[]{
  \includegraphics[trim=5mm 0mm -5mm 0mm, width=0.93in]{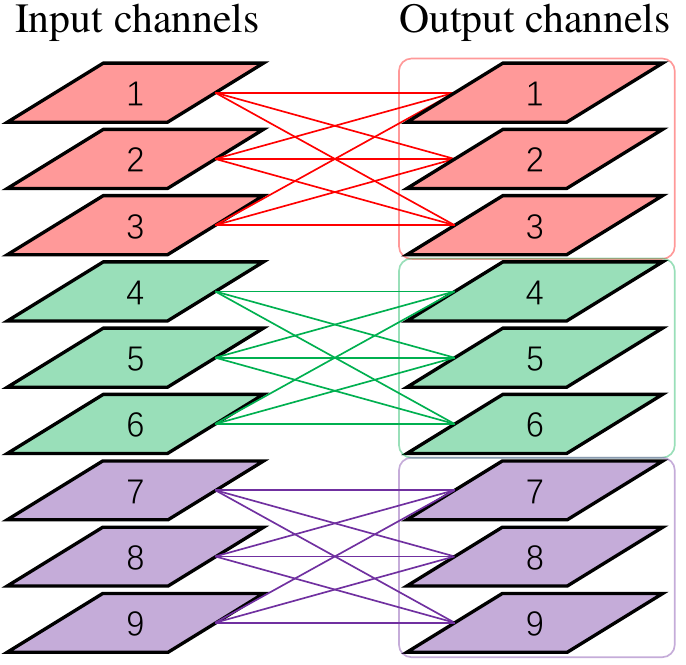}}
  \hspace{0in}
  \subfigure[]{
  \includegraphics[trim=5mm 0mm -5mm 0mm, width=0.93in]{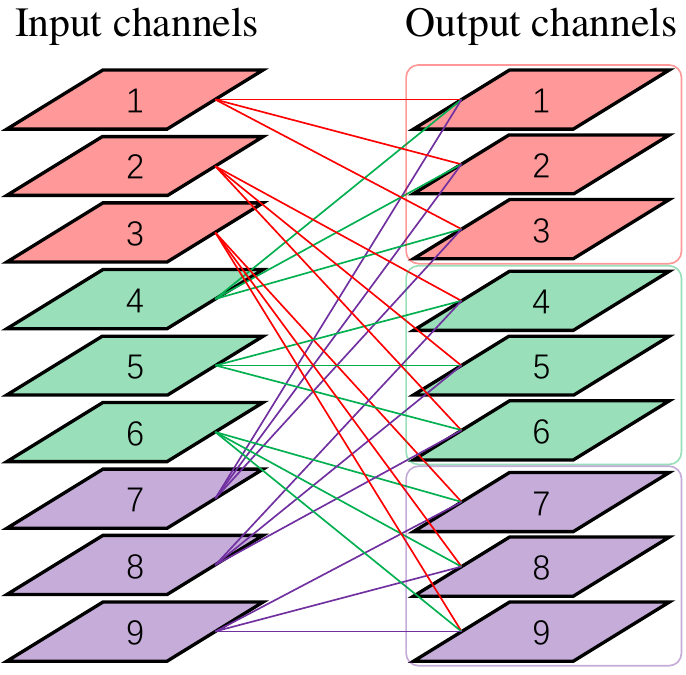}}
  \hspace{0in}
  \subfigure[]{
  \includegraphics[trim=5mm 0mm -5mm 0mm, width=0.93in]{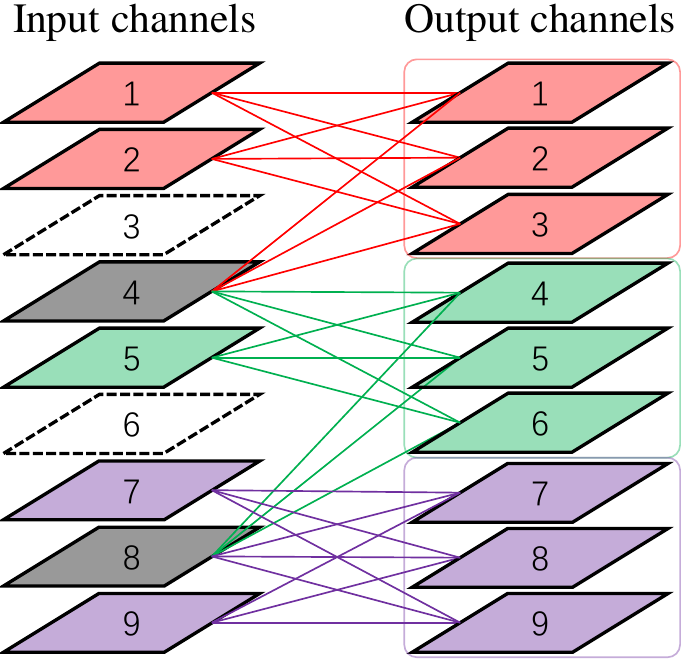}}
  \hspace{0in}
  \subfigure[]{
  \includegraphics[trim=5mm 0mm -5mm 0mm, width=0.93in]{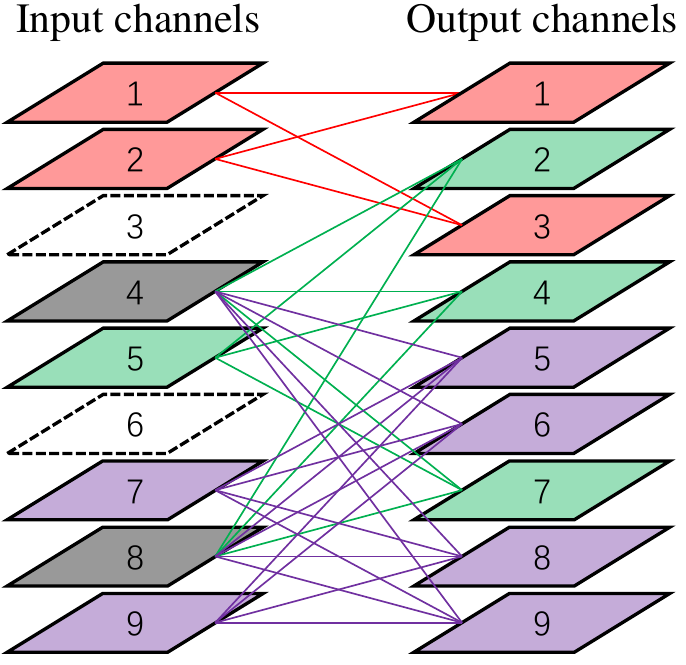}}
  {\caption{Evolution of group convolutions. (a) Regular convolution. (b) Regular group convolution. (c) Permuting group convolution. (d) Learned group convolution. (d) Self-grouping convolution. Note that white channels represent the ignored input channels, and gray channels indicate the reused input channels.}
  \label{fig:1}}
\end{minipage}
\end{figure*}

Recently, an enormous progress has been made in deep neural networks in connection with various computer vision tasks, such as image classification~\cite{KrizhevskySH12,SimonyanZ14,ZhangZH015}, object detection~\cite{GirshickDDM13,Girshick15,RenHGS15}, semantic segmentation~\cite{LongSD14,RonnebergerFB15,ZhouSTL18} and visual tracking~\cite{XuFWK18}, etc. Increasingly deeper network architectures are designed to improve performance, by optimising a huge set of parameters, involving heavy computation. However, most embedded systems and mobile platforms cannot afford such huge memory requirements and intensive computation due to their constrained resources~\cite{GuoZLKS18}. This severely impedes the application of deep neural networks. Lots of evidence has been provided to show that deep neural networks tend to be over-parameterised, and can be compressed with little, or no loss of accuracy. Many methods have been proposed to compress and accelerate deep neural networks, including pruning methods~\cite{HanPTD15,HuPTT16,WenWWCL16,LuoZZXWL18}, quantization methods~\cite{CourbariauxBD15,RastegariORF16,LiL16,ZhuHMD16}, decomposition with low rank~\cite{MasanaWHBA17,PengTLZXP18}, and designing compact architectures~\cite{IandolaMAHDK16,HowardZCKWWAA17,ZhangZLS17,HuangLW16}.

The key processing step in convolutional neural networks is convolution, in which each output channel corresponds to one filter over all of the input channels. Different from regular convolution, group convolution separately divides the input channels into multiple disjoint filter groups, thus convolutions are independently performed within each group for the reduction of computation budget and parameter cost. Since group convolution has an efficiently compact structure, and is particularly suitable for mobile and embedded applications, it has been attracting increasing interest as a means  to compress and accelerate deep neural networks.  These two convolutional architectures are illustrated in Fig.~\ref{fig:1} (a) and (b), respectively. The group convolution was first used in AlexNet~\cite{KrizhevskySH12} to handle the shortage of GPU's memory and surprisingly it delivered remarkable performance in image classification on ImageNet. Inspired by this idea,~\cite{XieGDTH16} constructed an efficient architecture, named ResNeXt, by combining a stacking strategy and a multi-branch architecture with group convolution, achieving a better classification result on ImageNet than its ResNet counterpart at a lower  computational complexity.~\cite{ZhangQXW17} presented a novel modularized neural network built by stacking interleaved group convolution (IGC) blocks, composed of primary and secondary group convolutions. To improve the representational power, IGC permutes the output channels of primary group convolutions as input channels of secondary group convolutions. Similarly, ShuffleNet~\cite{ZhangZLS17} introduced an efficient architecture in which two operations of point-wise group convolution and channel shuffle are adopted to significantly reduce the computational complexity, without degrading classification accuracy. Based on a similar idea,~\cite{GaoWJ18} used a channel-wise convolution to perform information fusion for the features outputted by prior independent groups. These methods permute the output channels of each group and put them into all the groups of the subsequent convolutional layer, such that the features of different groups interact with each other in a predefined manner. This type of architecture,  shown in Fig.~\ref{fig:1} (c), is called  permuting group convolution.~\cite{HuangLMW18} proposed a learned group convolution, in which a compact network architecture, termed CondenseNet, is constructed using dense connectivity, as shown in Fig.~\ref{fig:1} (d). CondenseNet is distinguished from the above methods in that each input channel is incorporated into one filter group by learning, rather than being pre-determined. It exhibits a better computational efficiency than MobileNet~\cite{HowardZCKWWAA17} and ShuffleNet~\cite{ZhangZLS17} at the same level of accuracy.

The above methods aim at selecting input channels for each filter group to improve the performance of deep neural networks. However, they are constrained by predefined group structures. A fixed assignment of filters to independent groups is not conducive to enhancing the recognition capability of deep neural networks. Firstly, the initial filter grouping in predefined grouping designs is data-independent. Secondly, because of their simplicity, these group convolution architectures, in which each group has the same number of filters and input channels, are prevented from realising their potential representation capacity. We hypothesise that filter groups should not be homogeneous, but rather diverse in the spatial group size, so that the diversity of the architectural features of group convolution can exploit the representational potential of deep neural networks. PolyNet~\cite{ZhangLCL17} has verified that diverse structures can improve the performance of image recognition as an additional dimension of optimisation, beyond depth and width in network design.

In this paper, we propose a novel method of self-grouping convolutional neural networks, which automatically groups the filters for each convolutional layer by clustering, instead of being predefined, to compress and accelerate deep neural networks. A neural network guides each filter to learn different representations from its input information through training, and each input channel plays a different role for such representations. For each filter, we first evaluate the importance of its input channels by an importance vector. Each element of the importance vector conveys an importance value of the corresponding input channel. We then learn the filter groups by clustering the importance vectors, which is data-dependent. Considering the redundancy of parameters, a sparse structure of each filter group is realised by pruning their unimportant connections based on their cluster centroids. In this way, we convert regular convolutions into self-grouping convolutions, where the diversity of group structures is promoted by  differences in spacial group size. This distinguishes the proposed method  from existing group convolutions~\cite{KrizhevskySH12,ZhangZLS17,ZhangQXW17,HuangLMW18}. Subsequently, we compensate the accuracy loss from pruning by two fine-tuning schemes, namely  (1) global only fine-tuning and (2) both local and global fine-tuning. The computational complexity of  the resulting efficient and compact self-grouping convolutional neural network and its memory requirements are  further reduced by extending the proposed  self-grouping approach to the fully-connected layers.

In Fig.~\ref{fig:1}, we illustrate the evolution of group convolutions, from regular group convolution, through permuting group convolution, learned group convolution, to our self-grouping convolution. By comprehensive experiments using various state-of-the-art CNN architectures, including ResNet~\cite{HeZRS16} and DenseNet~\cite{HuangLW16}, we show that our SG-CNN significantly reduces the size of network models and accelerates the inferences on popular vision datasets CIFAR-10/100~\cite{KrizhevskyH09} and ImageNet~\cite{ILSVRC15}, achieving superior performance. We present an ablation study that compares the performance of the proposed scheme in different conditions, which provides a deep insight into the properties of SG-CNNs. Furthermore, we also investigate the amenability  of our SG-CNN to generalisation by transfer learning, such as domain adaption and object detection.

The main contributions of this paper are summarized as follows:

\begin{itemize}
\item
A self-grouping convolution method for the compression and acceleration of deep neural networks by automatically converting regular convolutions into data dependent group convolutions with diverse group structures  that are learned using both, filter clustering, based on importance vectors, and network pruning based on cluster centroids.
\item
Our self-grouping method adapts to the fully-connected layers as well as the convolutional layers for extreme compression and acceleration.
\item
The proposed self-grouping method supports a  global only fine-tuning for an efficient network compression, preserving most of the information flow through data-dependent and diverse group structures.
\item
Comprehensive experiments testify that our self-grouping approach can be effectively applied to various state-of-the-art CNN architectures, including ResNet and DenseNet, with high compression ratio, low FLOPs and with a low, or no loss in accuracy.
\end{itemize}

The rest of this paper is organized as follows:
We first introduce the related work in Section~\ref{sec:RelatedWork}. We present our self-grouping convolution method in Section~\ref{sec:SG-CNNs}. Our self-grouping convolution is compared with previous group convolutions to elaborate its data-dependence and structural diversity by matrix decomposition in Section~\ref{sec:Analysis}. Subsequently, we validate our SG-CNN to show its superior performance through comprehensive experiments involving various network models and datasets in Section~\ref{sec:Experiments}. We present an ablation study, which enhances the understanding of SG-CNNs in Section~\ref{sec:AblationStudy}. We also investigate the generalization ability of SG-CNNs by transfer learning in Section~\ref{sec:GeneralizationAbility}. Finally, we draw the paper to conclusion in Section~\ref{sec:Conclusion}.


\section{Related Work}\label{sec:RelatedWork}

\textbf{Pruning methods.}
Pruning is one of the widely used methods to compress and accelerate deep neural networks. There are structural and non-structural pruning methods based on the sparsity of  spatial patterns.~\cite{HanPTD15} proposed a simple non-structural pruning strategy to compress deep neural networks by removing the connections corresponding to unimportant weights. Structural pruning methods have received considerable attention because they are a very direct way to obtain structurally sparse architectures.~\cite{MaoHPLLWD17} explored the effect of different pruning granularities on deep neural networks, and suggested coarse-grained pruning, such as connection-wise~\cite{HuangLMW18}, channel-wise~\cite{HeZS17}, filter-wise~\cite{LiKDSG16,LinJLDL19}, and even layer-wise~\cite{WenWWCL16} pruning, to compress and accelerate deep neural networks.~\cite{HeZS17} introduced a channel pruning method to compress deep neural networks. This method removes redundant channels through LASSO regularization, and reduces the error accumulated from pruning by minimizing the reconstruction error at the output feature maps.~\cite{LiKDSG16} estimates the importance of each filter according to the absolute sum of their kernel weights, and removes the unimportant filters based on a threshold, implying that the filters with low magnitude weights tend to yield weak feature maps. Recently,~\cite{LinJLDL19} proposed a very effective method of pruning by structured sparsity regularization, achieving superior performance in terms of accuracy and speedup. In CondenseNet~\cite{HuangLMW18}, less important connections were removed from filter groups to directly get structurally sparse patterns during the condensing stage. In our paper, we also adopt connection-wise pruning method to design structurally  sparse architectures for filter groups.

\textbf{Designing compact architectures}
The motivation for applying deep neural networks on devices with constrained resources also encourages the studies of designing efficient and compact network architectures. AlexNet~\cite{KrizhevskySH12} was a pioneering study in designing a group convolution architecture, although the main motivation for its design was to address the shortage of GPU resources. ResNeXt~\cite{XieGDTH16} applied group convolutions in its building blocks to reduce the computation complexity and the number of parameters.~\cite{ZhangQXW17} proposed an interleaved group convolutional neural network (IGCNet) in which each building block consists of two separate group convolution layers. To enhance the representation power of building blocks, the input channels of secondary group convolutions are linked to each primary group convolution. Similar to~\cite{ZhangQXW17},~\cite{ZhangZLS17} introduced a channel shuffle operation for multiple group convolutions to improve the representation power. These  methods exhibited recognition accuracy comparable to that of the original network, while  achieving low computational complexity. But they have one drawback in common, that is, the composition of input channels as well as the output channels in each group is predetermined rather than learned.~\cite{HuangLMW18} recently presented a learned group convolution, in which input channels are learned for each group. However, the filter partitions are still predefined. Moreover, only $1\times1$ convolution groups are learned, excluding $3\times3$ convolutional and fully-connected layers. In contrast, our self-grouping method can be applied to all of these layers. Recently, ~\cite{WangKSC19} proposed a fully learnable group convolution (FLGC) method to dynamically optimize the grouping structure, focusing on the convolutional layers for acceleration while achieving better accuracy than CondenseNet. Additionally, although the group structure is fully learnable, binary selection matrices for input channels and filters are approximately optimized by applying a softmax function to confront the problem of performance degradation. Compared to \cite{WangKSC19}, our motivation is similar, but we automatically construct the grouping structure by clustering based on importance vectors and by pruning based on cluster centroids. What is more, our self-grouping approach can be applied not only to the convolutional layers but also the fully-connected layers for simultaneous compression and acceleration.

Depthwise separable convolution is also a significant building block, which consists of two separate layers~\cite{HowardZCKWWAA17,ZophVSL17,SandlerHZZC18,Chollet16}. The first layer is a depthwise convolution, which performs spatial filtering over each input channel, and it can be viewed as a special group convolution in which each filter group independently contains only one input channel. The other is called pointwise convolution which projects the output of the depthwise convolution into a new feature space by performing 1$\times$1 convolution over all of its input channels. Many state-of-the-art network architectures, such as MobileNet~\cite{HowardZCKWWAA17} and NASNet~\cite{ZophVSL17}, have adopted such a building block to tradeoff reasonable accuracy against model size. Moreover, in order to keep representational power, the non-linearity operation between the two layers is usually removed from depthwise separable convolution~\cite{SandlerHZZC18,Chollet16}. Recently,~\cite{GaoWJ18} proposed an efficient and compact channel-wise convolution which can be combined with group convolution and depth-wise separable convolution to achieve a better trade-off between efficiency and accuracy.


\section{SG-CNNs}\label{sec:SG-CNNs}

\begin{figure*}
\begin{minipage}{1\textwidth}
  \centering
  \subfigure[]{
  \includegraphics[trim=5mm 0mm -5mm 0mm, width=1.20in]{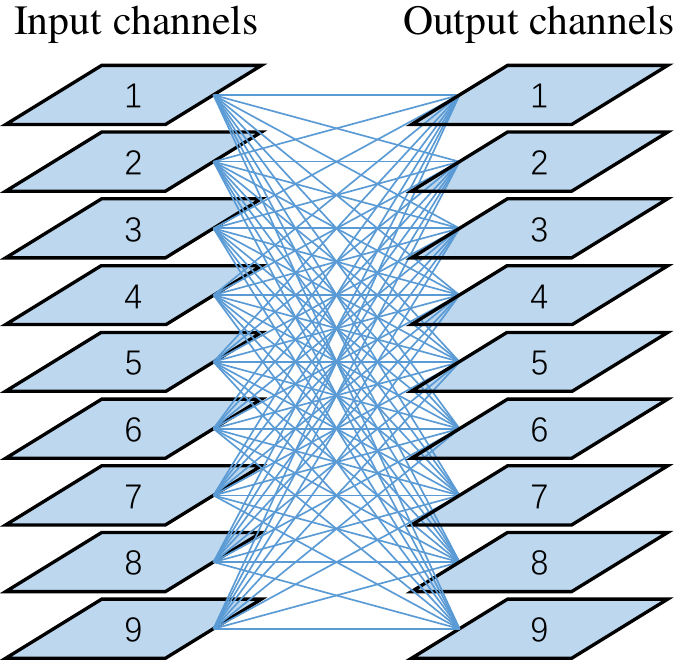}}
  \hspace{0in}
  \subfigure[]{
  \includegraphics[trim=5mm 0mm -5mm 0mm, width=1.20in]{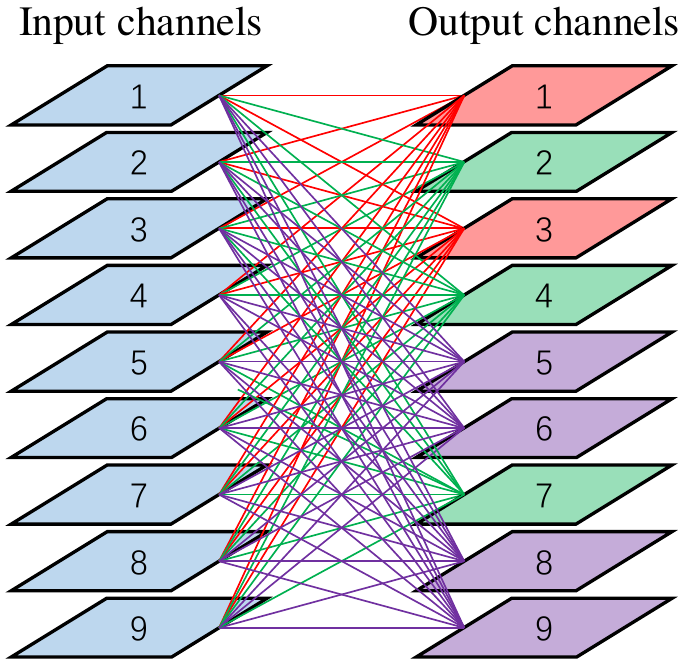}}
  \hspace{0in}
  \subfigure[]{
  \includegraphics[trim=5mm 0mm -5mm 0mm, width=1.20in]{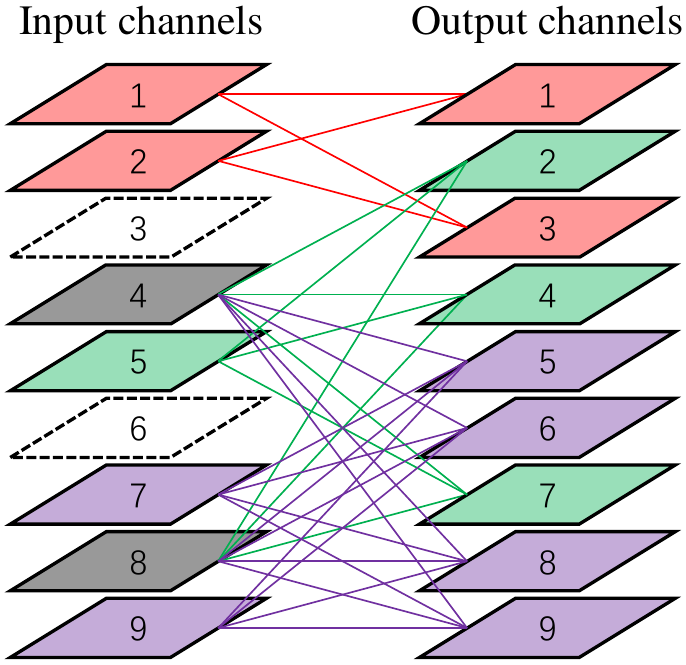}}
  \hspace{0in}
  \subfigure[]{
  \includegraphics[trim=5mm 0mm -5mm 0mm, width=1.20in]{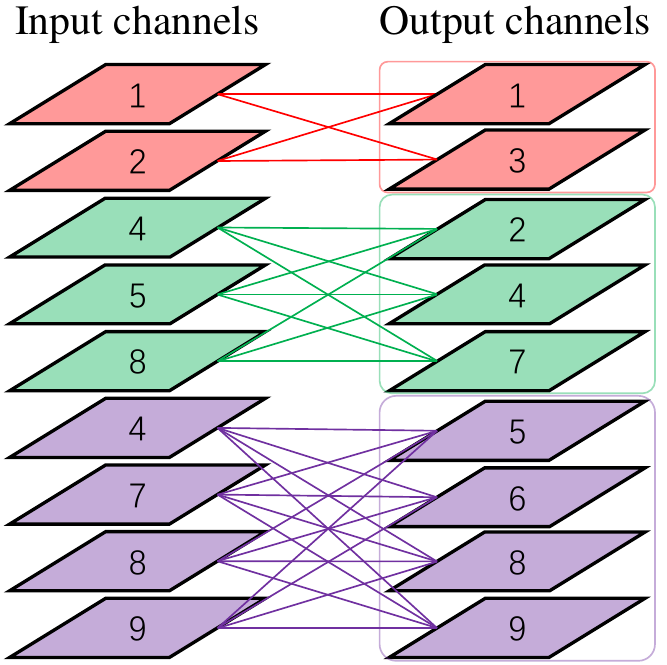}}
  {\caption{The overall pipeline of self-grouping convolutional neural networks. (a) Pretraining a regular convolution network. (b) Learning the filter groups by clustering based on importance vectors. (c) Learning the sparse structure for each filter group by a centroid-based pruning scheme. (d) Converting the sparsified convolution into regular group convolution with diverse group structures. Note that the same color filters represent that they have similar importance behaviors, white channels represent the ignored input channels, and gray channels indicate the reused input channels.
  }
  \label{fig:2}}
\end{minipage}
\end{figure*}

In this section, we first introduce the notation and preliminaries. Next, given a well-trained neural network, we introduce the concept of importance and the use of importance vectors in filter evaluation. Then, we present our self-grouping method to automatically cluster filters based on the similarity of their importance vectors.  A centroid-based pruning scheme is proposed to implement both the convolutional and fully-connected layers to compress and accelerate the neural network computation, followed by optional local fine-tuning and the obligatory global fine-tuning for the performance recovery. The outcome is a compact and efficient neural network with data-dependent and diverse group structures. We illustrate the overall pipeline of our self-grouping convolution in Fig.~\ref{fig:2}.

\subsection{Notations and Preliminaries}\label{sec:NotationsandPreliminaries}
Given an $L$-layer deep convolutional neural network, we denote the weights  of its $l$th convolutional layer as $\mathbf{W}\in\mathbb{R}^{C_{out}\times k\times k\times C_{in}}$, where $C_{out}$ and $C_{in}$ are the number of input channels and output channels, respectively, and $k$ is the kernel size. $\mathbf{x}\in\mathbb{R}^{k\times k\times C_{in}}$ is an input tensor which is obtained by sampling the input layer with $k\times k$ sliding window. Here, $\mathbf{W}$ and $\mathbf{x}$ can be viewed as a matrix with shape $C_{out}\times k\cdot k\cdot C_{in}$ and a vector with shape $k\cdot k\cdot C_{in}$, respectively, such that we have

\begin{equation}
\label{eqn:1}
\begin{aligned}
    \mathbf{y} = \mathbf{W}\mathbf{x},
\end{aligned}
\end{equation}

\noindent
where $\mathbf{y}\in\mathbb{R}^{C_{out}}$ is the corresponding output vector. $\mathbf{w}_{ij}\in\mathbf{W}$ corresponds to the $k\times k$ kernel of the $j$th input channel for the $i$th output one in the $l$th layer. For simplicity, we omit the bias term. In this paper, if not otherwise specified, all the notations indicate the parameters in the $l$th layer.

In order to reduce the computation cost and memory overhead, the regular group convolution approach focuses its convolution operation on the spatial or channel dimension of the filters. Suppose we partition $C_{out}$ filters and $C_{in}$ input channels into $g$ groups, denoted as $\mathbf{\dot{W}}_1, \mathbf{\dot{W}}_2, \ldots, \mathbf{\dot{W}}_g$, making each group to contain $C_{out}/g$ filters and $C_{in}/g$ input channels. Then the  regular group convolution can be formulated as follows,

\begin{equation}
\label{eqn:2}
\begin{aligned}
    \left[
    \begin{array}{cccc}
        \mathbf{\dot{y}}_{1}\\
        \mathbf{\dot{y}}_{2}\\
        \vdots\\
        \mathbf{\dot{y}}_{g}
    \end{array}
    \right]
    =
    \left[
    \begin{array}{cccc}
        \mathbf{\dot{W}}_1 & \mathbf{0}         & \ldots & \mathbf{0}\\
        \mathbf{0}         & \mathbf{\dot{W}}_2 & \ldots & \mathbf{0}\\
        \vdots             & \vdots             & \ddots & \mathbf{0}\\
        \mathbf{0}         & \mathbf{0}         & \ldots & \mathbf{\dot{W}}_g
    \end{array}
    \right]
    \left[
    \begin{array}{cccc}
        \mathbf{x}_{1}\\
        \mathbf{x}_{2}\\
        \vdots\\
        \mathbf{x}_{g}
    \end{array}
    \right],
\end{aligned}
\end{equation}

\noindent
where $\mathbf{x}_i\in \mathbb{R}^{k\cdot k\cdot C_{in}/g}$ is an input vector for group $i$, $\mathbf{\dot{y}}_i\in \mathbb{R}^{C_{out}/g}$ is the corresponding output vector of group $i$, and $\mathbf{\dot{W}}^i\in \mathbb{R}^{C_{out}/g\times k\cdot k\cdot C_{in}/g}$ denotes the weight block matrix of group $i$. Let $\mathbf{\dot{W}}$ = diag($\mathbf{\dot{W}}_1, \mathbf{\dot{W}}_2, \ldots, \mathbf{\dot{W}}_g$), which is a quasi-diagonal matrix,  assuming an equal group size, such that $\mathbf{\dot{y}}= \mathbf{\dot{W}}\mathbf{x}$.

For a fully-connected layer, we treat each of its neurons as convolutional channels with 1$\times$1 spatial size, i.e., $k$ = 1, such that we can obtain

\begin{equation}
\label{eqn:3}
\begin{aligned}
    \mathbf{y}^f = \mathbf{W}^f\mathbf{x}^f,
\end{aligned}
\end{equation}

\noindent
where $\mathbf{x}^f\in\mathbb{R}^{C_{in}}$ is an input vector,  $\mathbf{W}^f\in\mathbb{R}^{C_{out}\times C_{in}}$ is the weight matrix of the fully-connected layer, and $\mathbf{y}^f\in\mathbb{R}^{C_{out}}$ is the corresponding output vector. $w^f_{ij}\in\mathbf{W}^f$ is a scalar, and denotes the weight value of the $j$th input neuron for the $i$th output one. We also omit the bias term for simplicity.

By analogy, for the fully-connected layer, Eq. (\ref{eqn:3}) can be rewritten as $\mathbf{\dot{y}}^f= \mathbf{\dot{W}}^f\mathbf{x}^f$, i.e.,

\begin{equation}
\label{eqn:4}
\begin{aligned}
    \left[
    \begin{array}{cccc}
        \mathbf{\dot{y}}^f_1\\
        \mathbf{\dot{y}}^f_2\\
        \vdots\\
        \mathbf{\dot{y}}^f_g
    \end{array}
    \right]
    =
    \left[
    \begin{array}{cccc}
        \mathbf{\dot{W}}^f_1 & \mathbf{0}           & \ldots & \mathbf{0}\\
        \mathbf{0}           & \mathbf{\dot{W}}^f_2 & \ldots & \mathbf{0}\\
        \vdots               & \vdots               & \ddots & \mathbf{0}\\
        \mathbf{0}           & \mathbf{0}           & \ldots & \mathbf{\dot{W}}^f_g
    \end{array}
    \right]
    \left[
    \begin{array}{cccc}
        \mathbf{x}^f_1\\
        \mathbf{x}^f_2\\
        \vdots\\
        \mathbf{x}^f_g
    \end{array}
    \right],
\end{aligned}
\end{equation}

\noindent
where $\mathbf{x}^f_i\in \mathbb{R}^{C_{in}/g}$ is an input vector for group $i$, $\mathbf{\dot{y}}^f_i\in \mathbb{R}^{C_{out}/g}$ is the corresponding output vector of group $i$, and $\mathbf{\dot{W}}^f_i\in \mathbb{R}^{C_{out}/g\times C_{in}/g}$ denotes the block weight matrix of group $i$.

However, the limited spatial operation restricts the expressive power of the regular group convolution. To avoid this shortcoming, we propose a self-grouping convolution to relax the spatial restriction. This is achieved by clustering the filters based on the degree of similarity of the "so called" importance vectors and pruning the unimportant connections based on a centroid pruning strategy.

\subsection{Importance Vectors}\label{sec:EvaluatingImportanceVectors}
For a well-trained deep neural network shown in Fig.~\ref{fig:2} (a), its parameters are trained to make it attain a local or global optimum. Note that the training of neural networks effectively identifies the important parameters, while  inhibiting the less important connections. The distribution of these parameters conveys information about their relative importance. Generally, the parameters with low magnitudes tend to produce feature maps with weak activations, representing minor contributions to the neural network~\cite{HanPTD15,HuPTT16} output. On the contrary, the parameters of high magnitude are destined to make significant contributions. However, scalars cannot represent the information contained in a distribution. Considering group convolutions are closely relate to multiple filters and input channels, we introduce a novel concept, referred to as {\it importance vector}, for a filter to represent the importance of all its input channels.

For the $l$th layer, we define $\mathbf{V} = \{\mathbf{v}_1, \mathbf{v}_2, ..., \mathbf{v}_{C_{out}}\}$ as a set of the importance vectors of all its filters.  $\mathbf{v}_i$ corresponds to the $i$th filter, such that $\mathbf{v}_i = [v_{i1}~v_{i2}~...~v_{iC_{in}}]$ ($i$ = 1,2, ..., $C_{out}$), where $v_{ij}$ stands for the importance value of the $j$th input channel to the $i$th filter. We estimate $v_{ij}$ by the $\ell_1$-norm of its corresponding kernel $\mathbf{w}_{ij}$, as

\begin{equation}
\label{eqn:5}
\begin{aligned}
    v_{ij} = \left\|\mathbf{w}_{ij}\right\|_1.
\end{aligned}
\end{equation}

Similarly, for the fully-connected layers, we denote their importance vector set as $\mathbf{V}^f = \{\mathbf{v}^f_1, \mathbf{v}^f_2, ..., \mathbf{v}^f_{C_{out}}\}$, and $\mathbf{v}^f_i = [v^f_{i1}~v^f_{i2}~...~v^f_{iC_{in}}]$ ($i$ = 1,2, ..., $C_{out}$). The importance value $v^f_{ij}$ is estimated by the absolute value of its corresponding weight $w^f_{ij}$, as follows

\begin{equation}
\label{eqn:6}
\begin{aligned}
    v^f_{ij} = |w^f_{ij}|.
\end{aligned}
\end{equation}

Unlike the conventional methods in which the importance of these parameters is defined by scalars~\cite{HuPTT16,HuangLMW18,MolchanovTKAK16}, our method assesses their importance in terms of vectors. This concept suggests that the importance of weights should be gauged using the importance distribution of input channels for a filter. This can be achieved by assigning different filters into different groups by a clustering based on the similarity of the importance distributions.

\subsection{Self-grouping Filters by Clustering}\label{sec:Self-GroupingFiltersbyClustering}
In this part, we present how to automatically group filters by clustering based on the similarity of  importance vectors. Clustering is an efficient way to generate multiple filter groups where the behaviors of the input channels  is similar within each group but divergent between groups. Therefore, for the $l$th layer, we partition its importance vector set $\mathbf{V} = \{\mathbf{v}_1, \mathbf{v}_2, ..., \mathbf{v}_{C_{out}}\}$ into $g$ groups $\mathbf{G} = \{\mathbf{g}_1, \mathbf{g}_2, ..., \mathbf{g}_g\}$ by k-means clustering method so as to minimize the within-group sum of Euclidean distances, as follows,

\begin{equation}
\label{eqn:7}
\begin{aligned}
    \mathop{\arg\min}_{\mathbf{C}} \sum_{i=1}^{g} \sum_{\mathbf{v}_j \in \mathbf{g}_i} \left\|\mathbf{v}_j - \mathbf{c}_i\right\|_2.
\end{aligned}
\end{equation}

\noindent
Here, $\mathbf{C} = \{\mathbf{c}_1, \mathbf{c}_2, ..., \mathbf{c}_g\}$, and $\mathbf{c}_i = [c_{i1}~c_{i2}~...~c_{iC_{in}}]$ where $\mathbf{c}_i$ is the centroid vector of $\mathbf{g}_i$, and $c_{ij}$ corresponds to the $j$th input channels in $\mathbf{g}_i$. As shown in Fig.~\ref{fig:2} (b), all the filers are grouped into three groups for the convolutional layer, and each group has different group spatial size. Certainly, other clustering methods (e.g. k-medoids) could also be used for grouping the filters with similar importance vectors, which is beyond the scope of this paper.

Likewise, we apply the k-means clustering based on the similarity of  importance vectors to the fully-connected layers, thus achieving $g$ groups $\mathbf{G}^f = \{\mathbf{g}^f_1, \mathbf{g}^f_2, ..., \mathbf{g}^f_g\}$, satisfying the following condition,

\begin{equation}
\label{eqn:8}
\begin{aligned}
    \mathop{\arg\min}_{\mathbf{C}^f} \sum_{i=1}^{g} \sum_{\mathbf{v}^f_j \in \mathbf{g}^f_i} \left\|\mathbf{v}^f_j - \mathbf{c}^f_i\right\|_2,
\end{aligned}
\end{equation}

\noindent
where $\mathbf{C}^f = \{\mathbf{c}^f_1, \mathbf{c}^f_2, ..., \mathbf{c}^f_g\}$, and $\mathbf{c}^f_i$ stands for the centroid vector of $\mathbf{g}^f_i$, such that $\mathbf{c}^f_i = [c^f_{i1}~c^f_{i2}~...~c^f_{iC_{in}}]$. Here, $c^f_{ij}$ corresponds to the $j$th input neuron in $\mathbf{g}^f_i$.

The existing methods have aimed to design distinct group convolutions in which the filters are assigned to specific groups in a predefined manner and each group has the same number of filters, so that these designs are data-independent~\cite{XieGDTH16,ZhangQXW17,ZhangZLS17,GaoWJ18}. In contrast,  we automatically determine the filters for each group by clustering, instead of fixing a priori. Each group may have different number of filters, which is data-dependent. Therefore, self-grouping filters by clustering helps to enhance  the representation potential of group convolutions.

\subsection{Centroid-based Pruning Scheme}\label{sec:Centroid-based_Pruning_Scheme}
The requirement of group sparsity attracts increasing attention due to its beneficial effect on  compression and acceleration~\cite{WenWWCL16,AlvarezS16}. A connection based pruning can generate such structured sparse architecture for group convolutions by removing connections identified by negligible weights from groups. This enables parameter reduction and efficient computation~\cite{HanPTD15}. Furthermore, considering that the cluster centroids are representative importance vectors of their corresponding groups, we use them to determine the incoming input channels for each group. The result is  a centroid-based pruning scheme to construct our self-grouping convolution.

To be specific, we arrange each element of the centroid vectors in an ascending order to obtain a sorted set $\mathbf{I}$, as follows,

\begin{equation}
\label{eqn:9}
\begin{aligned}
    \mathbf{I} = \{c_{\widehat{1}}, c_{\widehat{2}}, ..., c_{\widehat{g\cdot C_{in}}}\}.
\end{aligned}
\end{equation}

\noindent
Here, $\widehat{i}$ indicates the order of $c_{\widehat{i}}$ in $\mathbf{I}$ to be $i$, and each element corresponds to multiple connections of its corresponding group. Then, we truncate the $n$ smallest values as follows,

\begin{equation}
\label{eqn:10}
\begin{aligned}
    top(\mathbf{I}, n) = \{c_{\widehat{1}}, c_{\widehat{2}}, ..., c_{\widehat{n}}\},
\end{aligned}
\end{equation}

\noindent
and prune their corresponding multiple weakest connections in the $l$th layer.

Correspondingly, for the fully-connected layers, the sorted set and the $n$ smallest values are defined as follows,

\begin{equation}
\label{eqn:11}
\begin{aligned}
    \mathbf{I}^f = \{c^f_{\widehat{1}}, c^f_{\widehat{2}}, ..., c^f_{\widehat{g\cdot C_{in}}}\},
\end{aligned}
\end{equation}

\begin{equation}
\label{eqn:12}
\begin{aligned}
    top(\mathbf{I}^f, n) = \{c^f_{\widehat{1}}, c^f_{\widehat{2}}, ..., c^f_{\widehat{n}}\}.
\end{aligned}
\end{equation}

Note that for a centroid vector, if some of its elements are within $top(\mathbf{I}, n)$, then partial connections of its corresponding whole group is discarded. As extreme cases, if all its elements are within $top(\mathbf{I}, n)$, then its corresponding whole group is discarded; On the contrary, if they are above $top(\mathbf{I}, n)$, then the corresponding whole group is reserved. As a consequence, different groups have different number of input channels. Moreover, the input channels can be shared by different groups, and can also be neglected by all the groups, which is similar to~\cite{HuangLMW18}.

In this way, the compression ratio of the $l$th layer can be calculated as follows,

\begin{equation}
\label{eqn:13}
\begin{aligned}
    r(\mathbf{G}, n) = \frac{\sum_{i=1}^{g}n_i \cdot \left\|\mathbf{g}_i\right\|}{\sum_{i=1}^{g}C_{in} \cdot \left\|\mathbf{g}_i\right\|},
\end{aligned}
\end{equation}

\noindent
where $\left\|\mathbf{g}_i\right\|$ denotes the number of filters in $\mathbf{g}_i$, and $n_i$ is the number of the $\mathbf{c}_i$'s elements that belong to $top(\mathbf{I}, n)$ in $\mathbf{g}_i$, such that $\sum_{i=1}^{g}n_i = n$. Further, the compression ratio of the neural network can be calculated as follows,

\begin{equation}
\label{eqn:14}
\begin{aligned}
    r = \frac{\sum_{l}\sum_{i=1}^{g}n_i \cdot \left\|\mathbf{g}_i\right\|}{\sum_{l}\sum_{i=1}^{g}C_{in} \cdot \left\|\mathbf{g}_i\right\|}.
\end{aligned}
\end{equation}

At each pruning iteration, the pruning step can be different, but for simplicity, the same pruning step is set for the $i$th layer to be $s$, which means the identical proportion  of connections are removed from the $i$th layer each time, which is closely related with $top(\mathbf{I}, n)$. In other words, after $t$ iterations, we truncate an appropriate number of $top(\mathbf{I}, n)$ from $\mathbf{I}$ to delete their corresponding connections, while satisfying the condition: $r(\mathbf{G}, n)\approx t\cdot s$.

So far, a self-grouping convolution with diverse structures has been formed by the remaining sparse connections. Such diverse structures significantly preserve the majority of information flow in each pruned layer, which helps to exploit the representation potential of group convolutions. The self-grouping convolution is shown in Fig.~\ref{fig:2} (c). Obviously, the connection pattern in self-grouping convolutions is controlled by $s$, $g$ and the training dataset together, where $g$ determines the number of filter groups. The filters of each group depend on the training dataset, and $s$ decides the number of input channels in each filter group. In Section ~\ref{sec:AblationStudy}, we investigate the effect of different $s$ and $g$ on the network performance to guide their setting in detail.

In summary, our self-grouping convolution method affords many advantages compared to the existing  pruning methods.: (1) By virtue of a novel centroid-based pruning scheme, we exploit the full knowledge  of weight parameter importance conveyed by the importance vector distribution. (2) Our proposed method preserves the majority of information flowing through the network, which helps achieving better recognition performance. (3) As our proposed method is appl;icable to the fully-connected layers as well as the convolutional layers, they can be pruned together for efficient compression and acceleration. (4) Different from the existing methods with a layer-by-layer grouping in a fixed manner, which impacts on the compression efficiency of networks with increasing depth, our method prunes the parameters for different layers in parallel. Therefore, it does not depend on the depth of the network but on the pruning step. This helps to improve the compression efficiency, especially for deep neural networks.

\subsection{Fine-tuning}\label{sec:Fine-tuning}
Although our proposed method minimises the performance degradation caused by the centroid-based pruning scheme, the cumulative error will damage the overall performance of the original neural networks. Therefore,  a fine-tuning that compensates  for the loss of accuracy from the pruning is desirable. There are two forms of fine-tuning: local fine-tuning and global fine-tuning. The former represents repeating local fine-tuning after each pruning to recover the performance of networks~\cite{HuPTT16,MolchanovTKAK16,LiuLSHYZ17}. This impacts on the computational time, while helping to maintain the network performance. The latter represents a global fine-tuning to strengthen the remaining part of the network to enhance its expressive ability~\cite{LinJLWHZ18}. Considering both, the performance and efficiency, we investigate two kinds of fine-tuning schemes: (1) global only fine-tuning and (2) both local and global fine-tuning. In section~\ref{sec:Experiments}, our extensive experiments on ImageNet testify that our self-grouping method obtains comparable results with each of  these two fine-tuning schemes, which convincingly shows our method preserves the majority of information flow through data-dependent and diverse group structures.

We depict the complete process of SG-CNNs to compress and accelerate deep network models in Algorithm.~\ref{alg:1}. Our self-grouping method prunes the unimportant connections from a well-trained neural network to reduce the size of the models and to accelerate the inference. The whole framework consists of five basic steps: (1) the importance vector computation for each filter; (2) filter grouping  by clustering based on their importance vectors; (3) prune unimportant connections based on the centroid-based pruning scheme; (4) (optionally) local fine-tuning the pruned networks; (5) global fine-tuning the pruned network.
%

\begin{algorithm}[htbp]
  \small
  \begin{algorithmic}[1]
    \REQUIRE ~~\\
        The well-trained neural network $\mathbf{N}$, the number $g$ of groups, the pruning step $s$, and the desired compression ratio $r_d$.
    \ENSURE ~~\\
        the compressed neural network $\mathbf{\hat{N}}$.
    \STATE $t$ = 1
    \REPEAT
        \FOR {each layer $l$ = 1 to $L$}
            \FOR {each filter $i$ = 1 to $C_{out}$}
                \STATE $\mathbf{v}_i = [v_{i1}~v_{i2}~...~v_{iC_{in}}]$, and $v_{ij} = \left\|\mathbf{w}_{ij}\right\|_1$
            \ENDFOR
            \STATE get $g$ groups $\mathbf{G} = \{\mathbf{g}_1, ..., \mathbf{g}_g\}$ and their cluster centroids $\mathbf{C} = \{\mathbf{c}_1, ..., \mathbf{c}_g\}$ by $\mathop{\arg\min}_{\mathbf{C}} \sum_{i=1}^{g} \sum_{\mathbf{v}_j \in \mathbf{g}_i} \left\|\mathbf{v}_j - \mathbf{c}_i\right\|_2$
            \STATE \ws{get $\mathbf{I}$ = $\{c_{\widehat{1}}, c_{\widehat{2}}, ..., c_{\widehat{g\cdot C_{in}}}\}$ by arranging each element of the centroid vectors in $\mathbf{C}$ in ascending order}
            \STATE pruning the weakest connections which belong to $top(\mathbf{I}, n)$ and satisfy $r(\mathbf{G}, n)\approx t\cdot s$
        \ENDFOR
        \IF {deploying local fine-tuning strategy}
            \STATE locally fine-tuning the pruned network $\mathbf{\hat{N}}$
        \ENDIF
        \STATE $t$ = $t$ + 1
    \UNTIL {$r\ge r_d$ }
    \STATE globally fine-tuning the pruned network $\mathbf{\hat{N}}$
  \end{algorithmic}
  \caption{Our self-grouping convolutional procedure.}
  \label{alg:1}
\end{algorithm}

\subsection{Deployment}\label{sec:Deploying}
When the compressed model is deployed on mobile devices or embedded platforms, we convert it into a network with regular connection patterns for inference speedups. Specifically, for each filter group, we duplicate the reused feature maps and delete the ignored feature maps. Afterwards, we rearrange these feature maps. The output channels are also rearranged to merge to locate the filters of the same group together. As a result, we obtain a regular group convolution with diverse group structures, which requires no special libraries or hardware for efficient inference, as shown in Fig.~\ref{fig:2} (d). The conversion process can easily be implemented by permutation matrices, as described in Section~\ref{sec:Analysis} in detail.


\section{Analysis}\label{sec:Analysis}

\begin{figure*}
\begin{minipage}{1\textwidth}
  \centering
  \subfigure[]{
  \includegraphics[trim=5mm 0mm -5mm 0mm, width=1.10in]{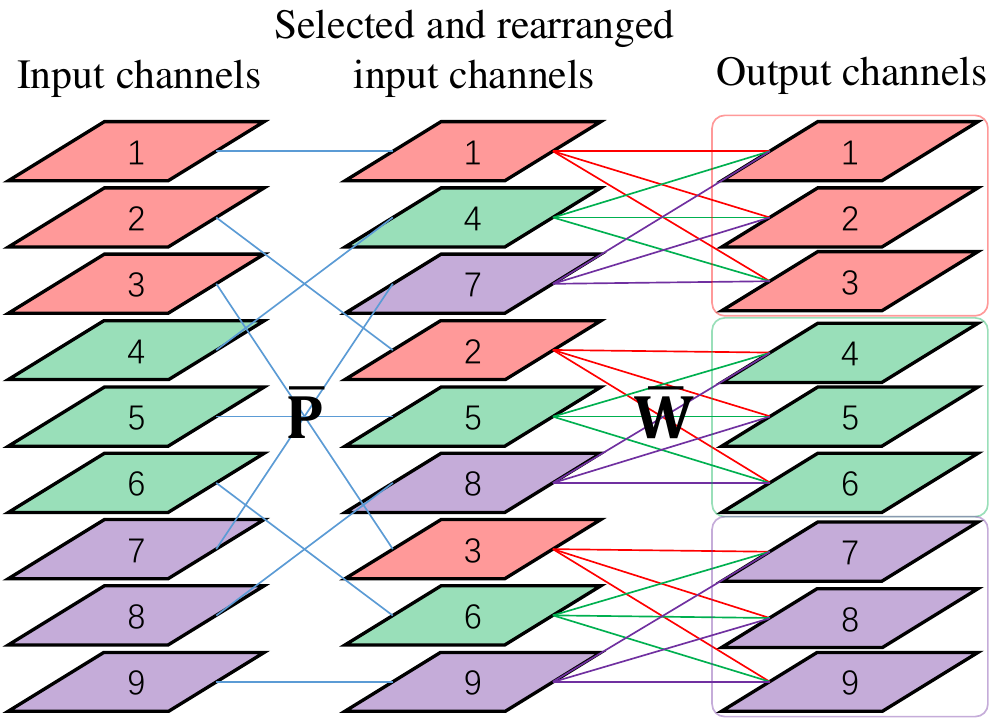}}
  \hspace{0in}
  \subfigure[]{
  \includegraphics[trim=5mm 0mm -5mm 0mm, width=1.10in]{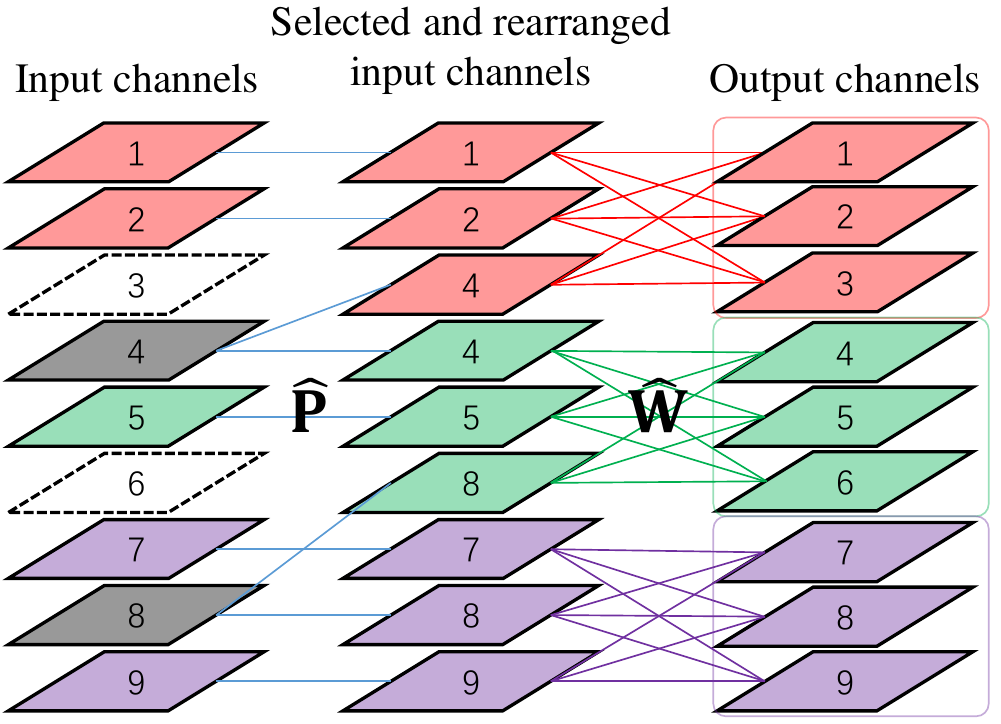}}
  \hspace{0in}
  \subfigure[]{
  \includegraphics[trim=5mm 0mm -5mm 0mm, width=1.50in]{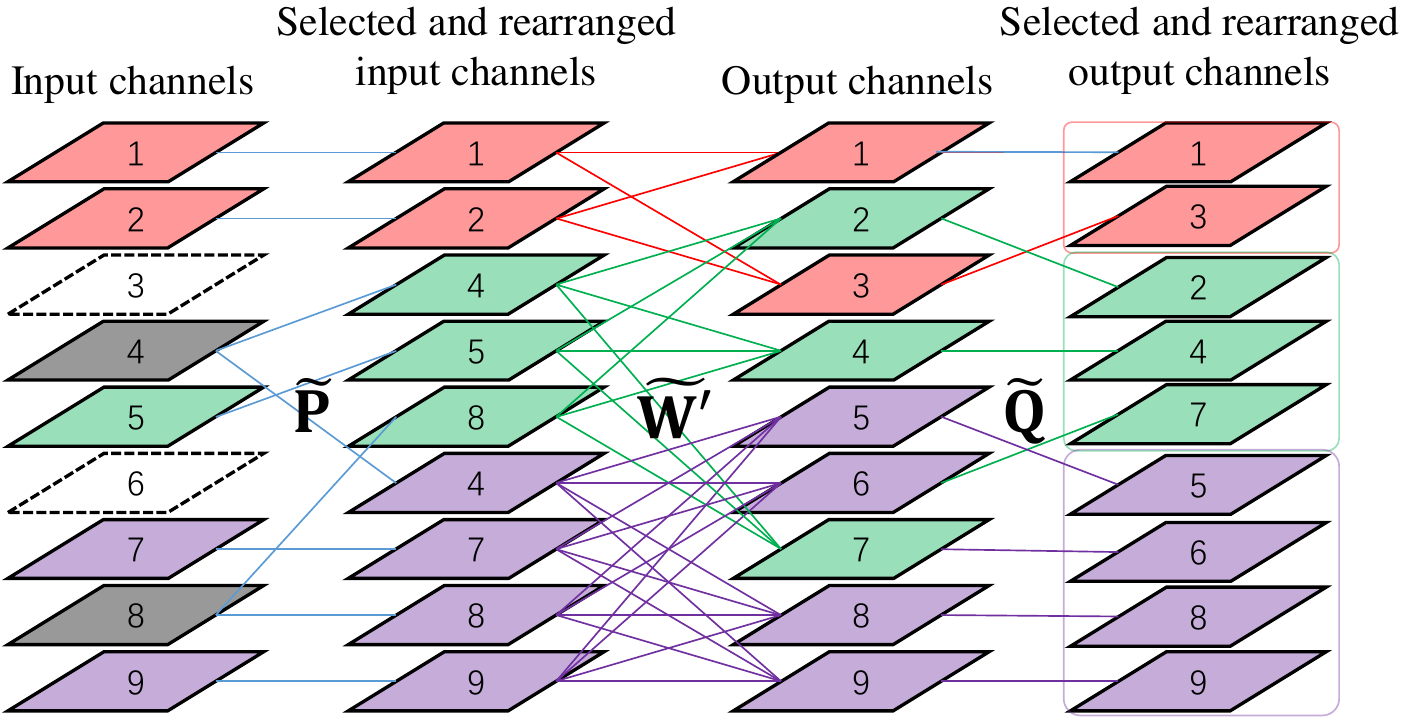}}
  \hspace{0in}
  \subfigure[]{
  \includegraphics[trim=5mm 0mm -5mm 0mm, width=1.10in]{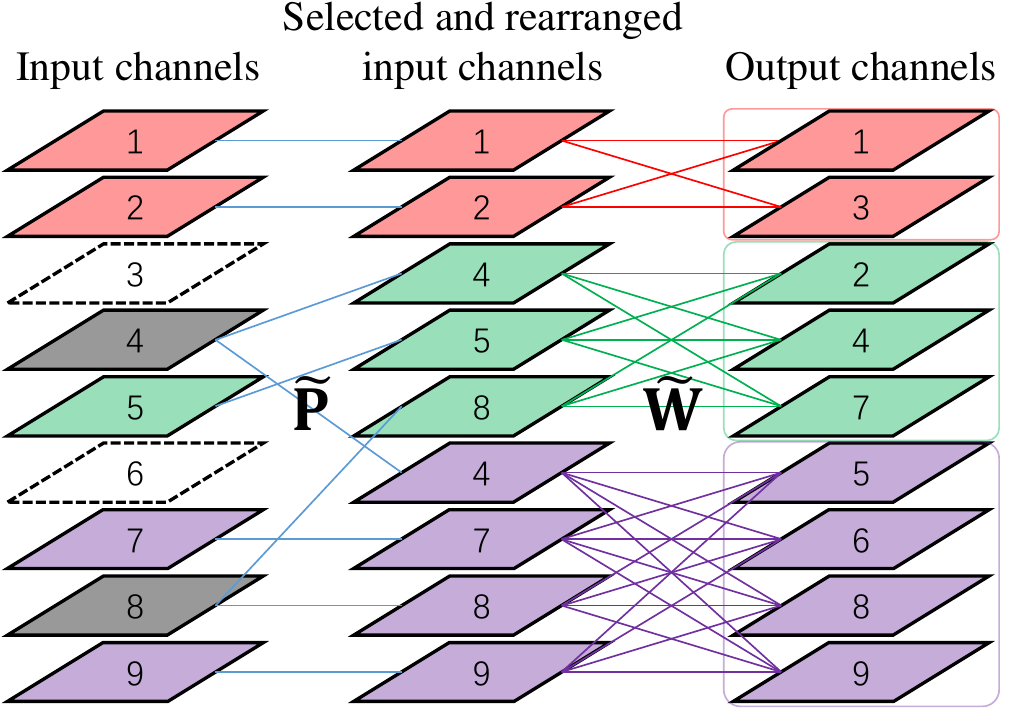}}
  {\caption{Matrix decomposition for different methods of group convolutions. (a) Permuting group convolution. (b) Learned group convolution. (c) Self-grouping convolution (I). (d) Self-grouping convolution (II).
  }
  \label{fig:4}}
\end{minipage}
\end{figure*}

Regular group convolution is highly restricted in its representational ability due to the limited scope of spatial calculations for each group. To enhance its representation power, a lot of methods have been introduced to relax the spatial restrictions, such as permuting output channels~\cite{ZhangQXW17}, shuffling channels~\cite{ZhangZLS17}, introducing channel-wise convolutions~\cite{GaoWJ18}, and using learned group convolutions~\cite{HuangLMW18}, which are equivalent to the deliberate selection of input channels for each disjoint group. However, they are rather simplistic in the composition of the filters for each groups. In the following, we compare our self-grouping convolution with these group convolutions to elaborate its data-dependence and structural diversity by matrix decomposition.

\noindent
\textbf{Permuting group convolution.}
For IGCNets, permuting the output channels of the primary group convolution can be interpreted as a specific selection of the input channels for each partition of the secondary group convolution, so that the input channels of the same secondary partition lie in different primary partitions~\cite{ZhangQXW17}. Similarly, shuffling channels can also be viewed as not only an organized rearrangement of input channels but also an intentional selection of input channels for each filter group to improve the representation capacity~\cite{ZhangZLS17}. The channel-wise convolution computes the output channels of each group from all input channels, while maintaining sparsity, which improves the interactions among filter groups for more representational power~\cite{GaoWJ18}.

The above networks have something in common. They have the same number of filters and input channels in each group, and a similar way to rearrange the input channels, as illustrated in Fig.~\ref{fig:4} (a). We formulate the permuting group convolution as follows,

\begin{equation}
\label{eqn:15}
\begin{aligned}
    \mathbf{\bar{y}} = \mathbf{\bar{W}}\mathbf{\bar{P}}\mathbf{x},
\end{aligned}
\end{equation}

\noindent
where $\mathbf{\bar{P}}$ is a permutation matrix to rearrange the order of input channels. It should be noted that $\mathbf{\bar{P}}$ is constant matrix due to predefined permutation designs. $\mathbf{\bar{W}}$ is a quasi-diagonal matrix, and the block structure of $\mathbf{\bar{W}}$ is also predefined. That is to say that sparse pattern of $\mathbf{\bar{P}}$ and $\mathbf{\bar{W}}$ is known before training.

\noindent
\textbf{Learned group convolution.}
By contrast with the above methods, learned group convolution also predefined the filters of each group, but learned input channels for each group based on its condensation criterion~\cite{HuangLMW18}. We show the equivalent group convolution in Fig.~\ref{fig:4} (b), and formulate it as follows,

\begin{equation}
\label{eqn:16}
\begin{aligned}
    \mathbf{\hat{y}} = \mathbf{\hat{W}}\mathbf{\hat{P}}\mathbf{x}.
\end{aligned}
\end{equation}

\noindent
Here, $\mathbf{\hat{P}}$ is a permutation matrix which is used to rearrange the input channels. Unlike $\mathbf{\bar{P}}$ in Equ. \ref{eqn:15}, $\mathbf{\hat{P}}$ is learnable to reuse the important input features and to ignore the less important ones. $\mathbf{\hat{W}}$ is the same as $\mathbf{\bar{W}}$ in the block structure which is predefined.

\noindent
\textbf{Self-grouping convolution.}
For our self-grouping convolution, the filters, as well as the input channels, cluster into different groups by learning. We split the filters in the same convolution layer into multiple groups by clustering, in contrast to prefixing. The input channels for each group are determined by centroid-based pruning.  In the convolution pattern, the number of the filters and input channels is different among groups. The input channels may be reused for different groups, and even may be ignored by all the groups, as shown in Fig.~\ref{fig:4} (c) and (d). Thus, we produce a diverse group convolutions with data-dependence, which is mathematically formulated as follows,

\begin{equation}
\label{eqn:17}
\begin{aligned}
    \mathbf{\tilde{y}} = (\mathbf{\tilde{Q}}\mathbf{\tilde{W}}')\mathbf{\tilde{P}}\mathbf{x} = \mathbf{\tilde{W}}\mathbf{\tilde{P}}\mathbf{x},
\end{aligned}
\end{equation}

\noindent
where both $\mathbf{\tilde{P}}$ and $\mathbf{\tilde{Q}}$ are permutation matrices, but different in function. $\mathbf{\tilde{P}}$ is used to rearrange the order of input channels, which is the same as $\mathbf{\hat{P}}$ in function. Distinguished from the other methods, we introduce a novel $\mathbf{\tilde{Q}}$ to organize the filters into multiple distinct groups, such that the sparse matrix $\mathbf{\tilde{W}}'$ is transformed into the block diagonal matrix $\mathbf{\tilde{W}}$. More importantly, these two permutation matrices are learned, rather than predefined, by clustering based on the similarity of importance vectors and pruning based on the cluster centroids. In contrast to $\mathbf{\dot{W}}$, $\mathbf{\bar{W}}$ and $\mathbf{\hat{W}}$ which have equal size blocks, $\mathbf{\tilde{W}}$ is a block diagonal matrix, but may have blocks of different size. The design is data-dependent because its block structure strongly depends on the training dataset. As a result, our self-grouping convolution operation is very effective and diverse, and does not require manually predefined permutation operations to improve the interaction of groups for better performance. This is verified by experiments in section~\ref{sec:Experiments}.


\section{Experiments}\label{sec:Experiments}

In this section, we empirically demonstrate the efficacy and efficiency of our proposed SG-CNN on four highly competitive computer vision recognition benchmark tasks, i.e., CIFAR-10/100~\cite{KrizhevskyH09} and ImageNet~\cite{ILSVRC15}. Comprehensive experiments are carried out on several state-of-the-art network architectures, including ResNet~\cite{HeZRS16} and DenseNet~\cite{HuangLW16}. All the experiments are implemented in pyTorch and are run on NVIDIA TITAN Xp GPU card with 12GB and 128G RAM. Actually, for simplicity, the same number of groups is set for each compressed layer. Additionally, there are few parameters (e.g. 3 channels for RGB images and 1 channel for gray images) in the first convolutional layer, but they are crucial as they provide the  original input information for the neural networks. Therefore, in order to keep enough input information, we do not compress the first convolutional layer.

\subsection{Datasets}\label{sec:Datasets}
\noindent
\textbf{CIFAR-10/100.} These two datasets consist of 50,000 images for training and 10,000 images for testing. The resolution of the images  is 32$\times$32. The data sets contain 10 and 100 categories, respectively. Due to the limited number of training samples, we augment the training datasets by random cropping and padding, and by horizontal flipping, which is the same technique of data augmentation adopted in~\cite{LiL16}.

\noindent
\textbf{ImageNet.} ILSVRC2012, a subset of the ImageNet dataset, contains 1.2M training images and 50K validation images as test samples. The image samples are categorized into 1000 classes. We follow the data augmentation scheme described in~\cite{HeZRS16}, i.e., each sample is randomly cropped to the 224$\times$224 size from the  rescaled   256$\times$256 size, and  horizontally flipped. We also apply a 224$\times$224 center crop to each test sample from the rescaled 256$\times$256 size at the test time.

\subsection{DenseNet on CIFAR-10/100}\label{sec:DenseNetonCIFAR-10/CIFAR100}
\noindent
\textbf{Model.}
For CIFAR-10/100, we use two modified version of DenseNet121 as our baselines, and train them with the same hyper-parameters for 200 epochs from scratch. The batch-size is set to be 64, weight decay 1e-4 and momentum 0.9. We follow the learning rate schedule: 0.1 for the first 100 epochs, 0.01 until epoch 150, and 0.001 to epoch 200. Finally, we obtain the baseline models with 95.23\% top-1 accuracy and 99.86\% top-5 accuracy for CIFAR-10 and 78.67\% top-1 accuracy and 94.55\% top-5 accuracy for CIFAR-100.

\noindent
\textbf{Implementation.}
For both DenseNet121 models on CIFAR-10/100, the number of groups is set to 8 for each layer. We simultaneously prune both the convolutional and fully-connected layers, and 5\% parameters are discarded from these layers each time. After each pruning, we locally fine-tune the pruned network for 4 epochs with a constant learning rate of 0.001. Finally, we globally tune the pruned networks for 200 epochs with the same hyper-parameters as in training, i.e., batch-size, weight decay, momentum and learning rate decay schedule, except for the initial learning rate of 0.01.

\noindent
\textbf{Results.}
We report the compression result of DenseNet on CIFAR-10 in Table~\ref{tab:2}. When the compression ratio is not more than 85\%, our approach achieves higher recognition accuracy and lower FLOPs than the original network model.

Firstly, we compare our SG-CNN with several state-of-the-art group convolution methods  to demonstrate the efficacy of our method. Compared with IGC~\cite{ZhangQXW17,XieWZLHQ18,SunLLW18}, our SG-CNN achieves accuracy that is 0.37\% higher than the best result of IGC's three versions at the approximate model size (1.71M vs. 2.2M). Also, compared with CondenseNet~\cite{HuangLMW18}, our SG-CNN achieves comparative recognition performance at approximate model size (0.68M vs. 0.52M). It is slightly inferior to CondenseNet at almost the same model size (0.34M vs. 0.33M), while achieving about 30\% lower FLOPs than CondenseNet, which means our SG-CNN has faster inference speed than CondenseNet. As for FLGC~\cite{WangKSC19}, with fully learned group convolutions, our SG-CNN is better by up to 1.8\% at the same model size (0.68M vs. 0.68M).

Secondly, our SG-CNN is also compared with other pruning methods. We can see that our SG-CNN surpasses Slimming~\cite{LiuLSHYZ17} by 0.37\% and 0.22\% in top-1 accuracy at comparable model size (1.37M vs. 1.44M and 0.68M vs. 0.66M), while achieving lower FLOPs. Compared with DMRNet~\cite{ZhaoLMLZZTW18}, SG-CNN outperforms it by 0.36\% in top-1 accuracy at almost the same model size (1.71M vs. 1.7M). For Variational Pruning~\cite{ZhaoNZZZT19}, the gap reaches to 1.16\% in top-1 accuracy and more than 2$\times$ in FLOPs (0.34M vs. 0.42M). And then for root~\cite{IoannouRCC16}, our SG-CNN is better by a large margin in terms of FLOPs and top-1 accuracy.

Finally, it is worth highlighting that our SG-CNN even surpasses other counterparts constructed by shift operations~\cite{ChenXZP19,WHWYJZGGGK18,JeonK18}. At comparable model size (0.68M vs. 0.55M, 1.71M vs. 1.76M and 1.03M vs. 0.99M), our SG-CNN surpasses them by 1.23\%, 2.61\% and 0.86\% in top-1 accuracy, respectively, while achieving lower FLOPs.

\begin{table}[!t]
  \scriptsize
  \caption{A comparison of several state-of-the-art methods for DenseNet121 on CIFAR-10.}
  \label{tab:2}
  \centering
  \setlength{\tabcolsep}{1.8mm}{
  \begin{tabular}{l|rr|rr|r}
    \specialrule{0.10em}{0pt}{0pt}
    Model &
    \makecell*[c]{Params} &
    \makecell*[c]{FLOPs} &
    \makecell*[c]{Top-1\\ (\%)} &
    \makecell*[c]{Top-5\\ (\%)} &
    Epochs \\
    \specialrule{0.10em}{0pt}{0pt}
    Baseline (k = 32)                   & 6.89M     & 888.36M   & 95.23          & 99.86  & 200    \\
    DenseNet (Conv-75/FC-75)            & 1.71M     & 221.90M   & \textbf{95.40} & 99.91  & 200+15*4+200 \\
    DenseNet (Conv-80/FC-80)            & 1.37M     & 177.72M   & 95.29          & 99.91  & 200+16*4+200 \\
    DenseNet (Conv-85/FC-85)            & 1.03M     & 134.10M   & 95.39          & 99.90  & 200+17*4+200 \\
    DenseNet (Conv-90/FC-90)            & 0.68M     &  89.77M   & 95.03          & 99.93  & 200+18*4+200 \\
    DenseNet (Conv-95/FC-95)            & 0.34M     &  45.76M   & 94.32          & 99.89  & 200+19*4+200 \\
    \specialrule{0.08em}{0pt}{0.5pt}
    \specialrule{0.08em}{0.5pt}{0pt}
    IGC-L4M8~\cite{ZhangQXW17}          & 0.96M     &  145M     & 90.12          & -      & 400    \\
    IGC-L4M8~\cite{ZhangQXW17}          & 0.57M     & 86.2M     & 92.81          & -      & 400    \\
    \specialrule{0.08em}{0pt}{0pt}
    IGC-L24M2~\cite{ZhangQXW17}         & 0.52M     & 94.8M     & 90.88          & -      & 400    \\
    IGC-L24M2~\cite{ZhangQXW17}         & 0.31M     & 57.1M     & 92.86          & -      & 400    \\
    \specialrule{0.08em}{0pt}{0pt}
    IGCV2*-C416~\cite{XieWZLHQ18}       & 0.65M     & -         & 94.51          & -      & 400    \\
    \specialrule{0.08em}{0pt}{0pt}
    IGCV3~\cite{SunLLW18}               &  2.2M     & -         & 95.03          & -      & 400    \\
    \specialrule{0.08em}{0pt}{0.5pt}
    \specialrule{0.08em}{0.5pt}{0pt}
    CondenseNet~\cite{HuangLMW18}       & 0.52M     & 122M      & 95             & -      & 300    \\
    CondenseNet~\cite{HuangLMW18}       & 0.33M     &  65M      & 95             & -      & 300    \\
    \specialrule{0.08em}{0pt}{0.5pt}
    \specialrule{0.08em}{0.5pt}{0pt}
    ResNet50-FLGC2~\cite{WangKSC19}     & 0.68M     &  44M      & 93.23          & -      & -    \\
    ResNet50-FLGC1~\cite{WangKSC19}     & 0.22M     &  23M      & 92.05          & -      & -    \\
    \specialrule{0.08em}{0pt}{0pt}
    MobileNetV2-FLGC(G=2)~\cite{WangKSC19} & 1.18M  & 158M      & 94.11          & -      & -    \\
    MobileNetV2-FLGC(G=3)~\cite{WangKSC19} & 0.85M  & 122M      & 94.20          & -      & -    \\
    MobileNetV2-FLGC(G=4)~\cite{WangKSC19} & 0.68M  & 103M      & 94.16          & -      & -    \\
    MobileNetV2-FLGC(G=8)~\cite{WangKSC19} & 0.43M  &  76M      & 93.09          & -      & -    \\
    \specialrule{0.08em}{0pt}{0.5pt}
    \specialrule{0.08em}{0.5pt}{0pt}
    ResNet-Slimming~\cite{LiuLSHYZ17}      & 1.44M  & 381M      & 94.92          & -      & 160+160    \\
    \specialrule{0.08em}{0pt}{0pt}
    DenseNet-Slimming~\cite{LiuLSHYZ17}    & 0.66M  & 381M      & 94.81          & -      & 160+160    \\
    \specialrule{0.08em}{0pt}{0.5pt}
    \specialrule{0.08em}{0.5pt}{0pt}
    DMRNet~\cite{ZhaoLMLZZTW18}            &  1.7M  & -         & 95.04          & -      & -    \\
    \specialrule{0.08em}{0pt}{0.5pt}
    \specialrule{0.08em}{0.5pt}{0pt}
    DenseNet-40 Pruned~\cite{ZhaoNZZZT19}  & 0.42M  & 156M      & 93.16          & -      & 300    \\
    \specialrule{0.08em}{0pt}{0.5pt}
    \specialrule{0.08em}{0.5pt}{0pt}
    root-2~\cite{IoannouRCC16}             & 1.64M  & 737M      & 92.09          & -      & -    \\
    root-4~\cite{IoannouRCC16}             & 1.23M  & 455M      & 92.02          & -      & -    \\
    root-8~\cite{IoannouRCC16}             & 1.03M  & 315M      & 92.15          & -      & -    \\
    root-16~\cite{IoannouRCC16}            & 0.93M  & 245M      & 91.67          & -      & -    \\
    \specialrule{0.08em}{0pt}{0.5pt}
    \specialrule{0.08em}{0.5pt}{0pt}
    ShiftResNet (SSL)~\cite{ChenXZP19}     & 0.55M  & 166M      & 93.8           & -      & -    \\
    \specialrule{0.08em}{0pt}{0.5pt}
    \specialrule{0.08em}{0.5pt}{0pt}
    ShiftResNet~\cite{WHWYJZGGGK18}        & 1.76M  & 279M      & 92.79          & -      & -    \\
    ShiftResNet~\cite{WHWYJZGGGK18}        & 0.87M  & 151M      & 92.74          & -      & -    \\
    ShiftResNet~\cite{WHWYJZGGGK18}        & 0.28M  &  67M      & 91.69          & -      & -    \\
    \specialrule{0.08em}{0pt}{0.5pt}
    \specialrule{0.08em}{0.5pt}{0pt}
    ASNet~\cite{JeonK18}                   & 0.99M  & -         & 94.53          & -      & -    \\
    \specialrule{0.10em}{0pt}{0pt}
  \end{tabular}}
\end{table}

Table~\ref{tab:3} shows the compression results of DenseNet on CIFAR-100. From the results, we note that our SG-CNN achieves 0.11\% higher top-1 accuracy than the original network model at the compression ratio of 70\%. As the compression ratio increases, the network gradually degrades in recognition accuracy, while achieving lower and lower FLOPs.

Our proposed method is compared with existing methods of group convolution to show its effectiveness. Compared with IGC~\cite{ZhangQXW17,SunLLW18}, it is significantly better, demonstrating that  our self-grouping convolutions are more expressive. CondenseNet~\cite{HuangLMW18} is outperformed  by 0.37\% at comparable model size (0.71M vs. 0.52M). However, our method slightly underperforms at approximate model size (0.36M vs. 0.33M), while achieving about 1/4 lower FLOPs than CondenseNet.

Compared with Slimming~\cite{LiuLSHYZ17}, our method achieves over 1\% and over 2\% higher top-1 accuracy at circa  1$\times$ and 3$\times$ lower FLOPs at approximately equal model size (1.40M vs. 1.46M and 0.71M vs. 0.66M). Compared with DMRNet~\cite{ZhaoLMLZZTW18}, our SG-CNN achieves 2.81\% higher top-1 accuracy at almost the same model size (1.75M vs. 1.7M). For Variational Pruning~\cite{ZhaoNZZZT19}, our SG-CNN surpasses it by up to 4.54\% top-1 accuracy at comparable compression ratio (0.71M vs. 0.65M), while achieving about 1.5$\times$ lower FLOPs.

Finally, in contrast to the methods constructed by shift operations~\cite{ChenXZP19,WHWYJZGGGK18,JeonK18}, our SG-CNN is better by 4.33\%, 4.3\% and 1.45\% in top-1 accuracy at comparable model size (0.71M vs. 0.55M, 1.75M vs. 1.76M and 1.06M vs. 0.99M), while achieving lower FLOPs.

\ws{Based on our observation, we find that most of training time is dominated by local fine-tuning in the experiments on both CIFAR-10 and CIFAR-100, which has a negative impact on the training efficiency. To this end, we try the other strategy, i.e., global only fine-tuning strategy, in the following experiments on ImageNet, and show the effectiveness and efficiency of our self-grouping convolution method by comparing between them.}

\begin{table}[!t]
  \scriptsize
  \caption{A comparison of several state-of-the-art methods for DenseNet121 on CIFAR-100.}
  \label{tab:3}
  \centering
  \setlength{\tabcolsep}{1.8mm}{
  \begin{tabular}{l|rr|rr|r}
    \specialrule{0.10em}{0pt}{0pt}
    Model &
    \makecell*[c]{Params} &
    \makecell*[c]{FLOPs} &
    \makecell*[c]{Top-1\\ (\%)} &
    \makecell*[c]{Top-5\\ (\%)} &
    Epochs \\
    \specialrule{0.10em}{0pt}{0pt}
    Baseline (k = 32)                   & 6.99M     & 888.45M   & 78.67          & 94.55  & 200    \\
    DenseNet (Conv-70/FC-70)            & 2.10M     & 266.60M   & \textbf{78.78} & 94.51  & 200+14*4+200 \\
    DenseNet (Conv-75/FC-75)            & 1.75M     & 222.14M   & 78.40          & 94.19  & 200+15*4+200 \\
    DenseNet (Conv-80/FC-80)            & 1.40M     & 176.46M   & 78.24          & 94.28  & 200+16*4+200 \\
    DenseNet (Conv-85/FC-85)            & 1.06M     & 133.46M   & 78.18          & 94.34  & 200+17*4+200 \\
    DenseNet (Conv-90/FC-90)            & 0.71M     &  89.86M   & 76.73          & 94.04  & 200+18*4+200 \\
    DenseNet (Conv-95/FC-95)            & 0.36M     &  45.67M   & 74.37          & 93.33  & 200+19*4+200 \\
    \specialrule{0.08em}{0pt}{0.5pt}
    \specialrule{0.08em}{0.5pt}{0pt}
    IGC-L4M8~\cite{ZhangQXW17}          & 0.96M     &  145M     & 64.48          & -      & 400    \\
    IGC-L4M8~\cite{ZhangQXW17}          & 0.57M     & 86.2M     & 67.81          & -      & 400    \\
    \specialrule{0.08em}{0pt}{0pt}
    IGC-L24M2~\cite{ZhangQXW17}         & 0.52M     & 94.8M     & 66.59          & -      & 400    \\
    IGC-L24M2~\cite{ZhangQXW17}         & 0.31M     & 57.1M     & 70.32          & -      & 400    \\
    \specialrule{0.08em}{0pt}{0pt}
    IGCV3~\cite{SunLLW18}               &  2.2M     & -         & 78.34          & -      & 400    \\
    \specialrule{0.08em}{0pt}{0.5pt}
    \specialrule{0.08em}{0.5pt}{0pt}
    CondenseNet~\cite{HuangLMW18}       & 0.52M     & 122M      & 76.36          & -      & 300    \\
    CondenseNet~\cite{HuangLMW18}       & 0.33M     &  65M      & 75.92          & -      & 300    \\
    \specialrule{0.08em}{0pt}{0.5pt}
    \specialrule{0.08em}{0.5pt}{0pt}
    ResNet-Slimming~\cite{LiuLSHYZ17}   & 1.46M     & 333M      & 77.13          & -      & 160+160    \\
    \specialrule{0.08em}{0pt}{0pt}
    DenseNet-Slimming~\cite{LiuLSHYZ17} & 0.66M     & 371M      & 74.72          & -      & 160+160    \\
    \specialrule{0.08em}{0pt}{0.5pt}
    \specialrule{0.08em}{0.5pt}{0pt}
    DMRNet~\cite{ZhaoLMLZZTW18}         &  1.7M     & -         & 75.59          & -      & -    \\
    \specialrule{0.08em}{0pt}{0.5pt}
    \specialrule{0.08em}{0.5pt}{0pt}
    DenseNet-40 Pruned~\cite{ZhaoNZZZT19} & 0.65M   & 218M      & 72.19          & -      & 300    \\
    \specialrule{0.08em}{0pt}{0.5pt}
    \specialrule{0.08em}{0.5pt}{0pt}
    ShiftResNet (SSL)~\cite{ChenXZP19}  & 0.55M     & 166M      & 72.4           & -      & -    \\
    \specialrule{0.08em}{0pt}{0.5pt}
    \specialrule{0.08em}{0.5pt}{0pt}
    ShiftResNet~\cite{WHWYJZGGGK18}     & 1.76M     & 279M      & 74.10          & -      & -    \\
    ShiftResNet~\cite{WHWYJZGGGK18}     & 0.87M     & 151M      & 73.64          & -      & -    \\
    ShiftResNet~\cite{WHWYJZGGGK18}     & 0.28M     &  67M      & 69.82          & -      & -    \\
    \specialrule{0.08em}{0pt}{0.5pt}
    \specialrule{0.08em}{0.5pt}{0pt}
    ASNet~\cite{JeonK18}                & 0.99M     & -         & 76.73          & -      & -    \\
    \specialrule{0.10em}{0pt}{0pt}
  \end{tabular}}
\end{table}

\subsection{ResNet and DenseNet on ImageNet}\label{sec:ResNetandDenseNetonImageNet}

\noindent
\textbf{Model.}
In this experiment, we investigate the proposed SG-CNN on two state-of-the-art CNN architectures, i.e., ResNet50 and DenseNet201. For a fair comparison, we use their network models pre-trained on ImageNet as our baseline networks instead of ones trained from scratch.

\noindent
\textbf{Implementation.}
For ResNet50 and DenseNet201, we set the number of groups to 16. These two models are pruned with the compression step of 10\% to get a series of models of different size. Considering the difference between the convolutional and fully-connected layers in redundancy, the compression ratio ranges from 10\% to 80\% for the convolutional layers and from 10\% to 60\% for the fully-connected ones. We apply two fine-tuning schemes, i.e. the local and global fine-tuning and the global only fine-tuning, to verify the effectiveness and efficiency of our method. For the former scheme, the local fine-tuning is performed for a small number 20 of epochs after each pruning, the learning rate is set to be 0.0001, and kept constant. For the global fine-tuning in these two schemes, the learning rate is set to be 0.01, which is divided by 10 at 30 and 60 epoch, respectively, until 90 epochs. The other hyper-parameters are set up as follows: batch-size 128, weight decay 0.0001 and momentum 0.9.

\noindent
\textbf{Results.}
We illustrate the compression result of ResNet50 on ImageNet in Table~\ref{tab:4}. For the models with and without local fine-tuning, there is little loss at low compression ratio. The loss gradually increases at high compression ratio, but the loss of top-1 accuracy is less than 3\%. Additionally, there is a very little gap between the two fine-tuning strategies, namely less than 0.2\% in top-1 accuracy. This result manifests that our proposed method preserves the flow of relevant information after pruning, which  improves the network performance.

In order to show the efficacy of our method, our SG-CNN is compared with the  state-of-the-art CNNs. First, we compare our SG-CNN with other group convolution methods, such as IGCV1~\cite{ZhangQXW17} and root~\cite{IoannouRCC16}. The gap between our SG-CNN and IGCV1 reaches  4.6\% and 2.87\% in top-1 and top-5 accuracy at comparable model size (11.88M vs. 11.329M), respectively. For root, we observe that it is outperformed by our SG-CNN  by a significant margin in recognition performance, while achieving a smaller model size and a lower computation complexity.

Second, our SG-CNN is compared with other compression methods, including ThiNet~\cite{LuoZZXWL18}, SSR~\cite{LinJLDL19}, GDP~\cite{LinJLWHZ18} and LRDKT~\cite{LinJCTL18}. It outperforms ThiNet by more than 3\% and 5\% in top-1 accuracy at comparable model sizes (11.88M vs. 12.38M and 7.76M vs. 8.66M), respectively. In comparison with SSR, our SG-CNN achieves better recognition performance, smaller model size and lower FLOPs. Thus, it can be seen that our proposed method is superior to SSR. Compared to GDP, which focuses on accelerating deep convolutional neural networks, our proposed method achieves lower FLOPs than GDP, while achieving better accuracy. For LRDKT, our SG-CNN obtains comparable recognition performance at low compression ratio (9.83M vs. 9.8M), surpassing LRDKT by more than 4\% at high compression ratio (7.76M vs. 6.3M).

\begin{table}[!t]
  \scriptsize
  \caption{A comparison of several state-of-the-art methods for ResNet50 on ILSVRC2012. Here, "ResNet-G" and "ResNet-LG" indicates the recognition accuracy with the global only fine-tuning and with both the local and global fine-tuning, respectively.}
  \label{tab:4}
  \centering
  \setlength{\tabcolsep}{1.8mm}{
  \begin{tabular}{l|rr|rr|r}
    \specialrule{0.10em}{0pt}{0pt}
    Model &
    \makecell*[c]{Params} &
    \makecell*[c]{FLOPs} &
    \makecell*[c]{Top-1\\ (\%)} &
    \makecell*[c]{Top-5\\ (\%)} &
    Epochs \\
    \specialrule{0.10em}{0pt}{0pt}
    Baseline                            & 25.55M    & 4.09G     & 76.13          & 92.86          & 90 \\
    ResNet-G (Conv-60/FC-60)            & 11.88M    & 1.91G     & \textbf{75.20} & 92.55          & 90+90 \\
    ResNet-G (Conv-70/FC-60)            &  9.83M    & 1.55G     & 74.43          & 92.30          & 90+90 \\
    ResNet-G (Conv-80/FC-60)            &  7.76M    & 1.20G     & 73.22          & 91.70          & 90+90 \\
    \specialrule{0.08em}{0pt}{0pt}
    ResNet-LG (Conv-60/FC-60)           & 11.87M    & 1.91G     & 75.12          & \textbf{92.59} & 90+6*20+90 \\
    ResNet-LG (Conv-70/FC-60)           &  9.83M    & 1.56G     & 74.42          & 92.31          & 90+7*20+90 \\
    ResNet-LG (Conv-80/FC-60)           &  7.76M    & 1.20G     & 73.38          & 91.69          & 90+8*20+90 \\
    \specialrule{0.08em}{0pt}{0.5pt}
    \specialrule{0.08em}{0.5pt}{0pt}
    IGCV1~\cite{ZhangQXW17}             & 11.329M   & 2.2G      & 70.6           & 89.68          & 95    \\
    IGCV1~\cite{ZhangQXW17}             & 11.205M   & 1.9G      & 69.23          & 89.01          & 95    \\
    IGCV1~\cite{ZhangQXW17}             &   8.61M   & 1.3G      & 73.05          & 91.08          & 95    \\
    \specialrule{0.08em}{0pt}{0.5pt}
    \specialrule{0.08em}{0.5pt}{0pt}
    root-2~\cite{IoannouRCC16}          & 25.4M     & 3.86G     & 72.7           & 91.2           & -    \\
    root-4~\cite{IoannouRCC16}          & 25.1M     & 3.37G     & 73.4           & 91.8           & -    \\
    root-8~\cite{IoannouRCC16}          & 23.2M     & 2.86G     & 73.4           & 91.8           & -    \\
    root-16~\cite{IoannouRCC16}         & 18.7M     & 2.43G     & 73.2           & 91.8           & -    \\
    root-32~\cite{IoannouRCC16}         & 16.4M     & 2.22G     & 72.9           & 91.5           & -    \\
    root-64~\cite{IoannouRCC16}         & 15.3M     & 2.11G     & 73.2           & 91.5           & -    \\
    \specialrule{0.08em}{0pt}{0.5pt}
    \specialrule{0.08em}{0.5pt}{0pt}
    ThiNet~\cite{LuoZZXWL18}            & 16.94M    & 2.44G     & 74.03         & 92.11           & 196+48    \\
    ThiNet~\cite{LuoZZXWL18}            & 12.38M    & 1.70G     & 72.03         & 90.99           & 196+48    \\
    ThiNet~\cite{LuoZZXWL18}            &  8.66M    & 1.10G     & 68.17         & 88.86           & 196+48    \\
    \specialrule{0.08em}{0pt}{0.5pt}
    \specialrule{0.08em}{0.5pt}{0pt}
    SSR-L2,1~\cite{LinJLDL19}           & 15.9M     & 1.9G      & 72.13         & 90.57           & 90+30    \\
    SSR-L2,0~\cite{LinJLDL19}           & 15.5M     & 1.9G      & 72.29         & 90.73           & 90+30    \\
    \specialrule{0.08em}{0pt}{0.5pt}
    \specialrule{0.08em}{0.5pt}{0pt}
    GDP~\cite{LinJLWHZ18}               & -         & 2.24G     & 72.61         & 91.05           & 90+20    \\
    GDP~\cite{LinJLWHZ18}               & -         & 1.88G     & 71.89         & 90.71           & 90+20    \\
    GDP~\cite{LinJLWHZ18}               & -         & 1.57G     & 70.93         & 90.14           & 90+20    \\
    \specialrule{0.08em}{0pt}{0.5pt}
    \specialrule{0.08em}{0.5pt}{0pt}
    LRDKT~\cite{LinJCTL18}              & 9.8M      & -         & 74.64         & 91.86           & 90+15    \\
    LRDKT~\cite{LinJCTL18}              & 6.3M      & -         & 69.07         & 88.5            & 90+15    \\
    \specialrule{0.10em}{0pt}{0pt}
  \end{tabular}}
\end{table}

For DenseNet201, we summarize the performance result on ImageNet in Table~\ref{tab:5}. For the models with and without local fine-tuning, their loss of top-1 accuracy is less than 2\%. The gap between them is less than 0.2\%, and the models without the local fine-tuning even achieve higher accuracy than those with the local fine-tuning at the same compression ratio. This result  verifies that our method preserves the relevant flow of information. Our SG-CNN achieves an acceleration of over 4 $\times$ FLOPs reduction for higher compression ratios.

First, we compare our method with several state-of-the-art of group convolution methods, showing outstanding performance in recognition accuracy. We compare our best results with two versions of ShuffleNet~\cite{ZhangZLS17,MaZZS18}, achieving 1.47\% and 1.31\% higher top-1 accuracy (4.32M vs. 5.3M and 6.00M vs. 7.4M), respectively. Additionally, two versions of MobileNet~\cite{HowardZCKWWAA17,SandlerHZZC18} are also compared with our best results. We observe that SG-CNN outperforms MobileNetV2 by about 1.5\% in top-1 accuracy (6.00M vs. 6.9M), and MobileNetV1 by up to 4.39\% in top-1 accuracy (4.32M vs. 4.2M). In contrast with SENet~\cite{HuSS18}, SG-CNN with a smaller model size achieves slightly higher top-1 accuracy (4.32M vs. 4.7M). We also compare our SG-CNN with IGCV2~\cite{XieWZLHQ18} and IGCV3~\cite{SunLLW18}. The gap reaches  4.47\% and 1.66\% in top-1 accuracy at comparable model size (4.32M vs. 4.1M and 6.00M vs. 7.2M), respectively. Compared to ChannelNet~\cite{GaoWJ18}, the largest gap is as much as 4.67\% in top-1 accuracy (4.32M vs. 3.7M). Compared to CondenseNet~\cite{HuangLMW18} and CondenseNet-FLGC~\cite{WangKSC19}, our method achieves 1.37\% and 0.47\% higher top-1 accuracy, and 0.9\% and 0.5\% higher top-5 accuracy for DenseNet-LG (4.32M vs. 4.8M). It obtains 1.19\% and 0.29\% higher top-1 accuracy and 0.85\% and 0.45\% higher top-5 accuracy for DenseNet-G (4.32M vs. 4.8M). But both CondenseNet and CondenseNet-FLGC obtain lower FLOPs than our SG-CNN, mainly benefiting from their deployment on a new dense architecture, which is instrumental in achieving a low computation complexity. Our self-grouping method outperforms these state-of-the-art group convolutions methods at similar compression ratios.

Compared to KSE~\cite{LiLZLDWHJ19}, our SG-CNN has a better performance by 1.27\% in top-1 accuracy and by 0.66\% in top-5 accuracy at approximately equal model size (4.32M vs. 4.21M), respectively. Moreover, at approximately equal computation complexity (0.99G vs. 0.9G), the gap reaches up to 2.14\% and 1.4\% in top-1 and top-5 accuracy, respectively.

Finally, our SG-CNN is compared with auto-searched networks, such as NASNet~\cite{ZophVSL17}, PNASNet~\cite{LiuZSHLFYHM17} and MnasNet~\cite{TanCPVL18}, which consume more time and GPUs to complete the search process. Clearly, our SG-CNN again achieves competitive recognition performance at the same model size.

\ws{We also observe that the global strategy significantly improves the training efficiency, and achieves good or even better accuracy than the strategy with local fine-tuning. These experimental results fully show that our self-grouping convolution can preserve considerable representation ability after pruning even without local fine-tuning.}

\begin{table}[!t]
  \scriptsize
  \caption{A comparison of several state-of-the-art methods for DenseNet201 on ILSVRC2012. Here, "DenseNet-G" and "DenseNet-LG" indicates the recognition accuracy with only global fine-tuning and with both local and global fine-tuning, respectively.}
  \label{tab:5}
  \centering
  \setlength{\tabcolsep}{1.8mm}{
  \begin{tabular}{l|rr|rr|r}
    \specialrule{0.10em}{0pt}{0pt}
    Model &
    \makecell*[c]{Params} &
    \makecell*[c]{FLOPs} &
    \makecell*[c]{Top-1\\ (\%)} &
    \makecell*[c]{Top-5\\ (\%)} &
    Epochs \\
    \specialrule{0.10em}{0pt}{0pt}
    Baseline                            & 19.82M    & 4.29G     & 76.90          & 93.37           & 90    \\
    DenseNet-G (Conv-70/FC-60)          &  6.00M    & 1.34G     & \textbf{76.21} & \textbf{93.07}  & 90+90 \\
    DenseNet-G (Conv-80/FC-60)          &  4.32M    & 0.99G     & 74.99          & 92.55           & 90+90 \\
    \specialrule{0.08em}{0pt}{0pt}
    DenseNet-LG (Conv-70/FC-60)         &  6.00M    & 1.34G     & 76.12          & 93.06           & 90+7*20+90 \\
    DenseNet-LG (Conv-80/FC-60)         &  4.32M    & 0.99G     & 75.17          & 92.60           & 90+8*20+90 \\
    \specialrule{0.08em}{0pt}{0.5pt}
    \specialrule{0.08em}{0.5pt}{0pt}
    ShuffleNetV1~\cite{ZhangZLS17}      & 5.3M      & 524M      & 73.7           & -               & 240    \\
    \specialrule{0.08em}{0pt}{0pt}
    ShuffleNetV2~\cite{MaZZS18}         & 7.4M      & 591M      & 74.9           & -               & 240    \\
    \specialrule{0.08em}{0pt}{0.5pt}
    \specialrule{0.08em}{0.5pt}{0pt}
    MobileNetV1~\cite{HowardZCKWWAA17}  & 4.2M      & 569M      & 70.6           & -               & -    \\
    \specialrule{0.08em}{0pt}{0pt}
    MobileNetV2~\cite{SandlerHZZC18}    & 6.9M      & 585M      & 74.7           & -               & -    \\
    \specialrule{0.08em}{0pt}{0.5pt}
    \specialrule{0.08em}{0.5pt}{0pt}
    SE-MobileNet~\cite{HuSS18}          & 4.7M      & 572M      & 74.7           & 92.1            & 100    \\
    SE-ShuffleNet~\cite{HuSS18}         & 2.4M      & 142M      & 68.3           & 88.3            & 100    \\
    \specialrule{0.08em}{0pt}{0.5pt}
    \specialrule{0.08em}{0.5pt}{0pt}
    IGCV2~\cite{XieWZLHQ18}             & 4.1M      & 564M      & 70.7           & -               & 100+20    \\
    IGCV2~\cite{XieWZLHQ18}             & 1.3M      & 156M      & 65.5           & -               & 100+20    \\
    IGCV2~\cite{XieWZLHQ18}             & 0.5M      & 46M       & 54.9           & -               & 100+20    \\
    \specialrule{0.08em}{0pt}{0pt}
    IGCV3~\cite{SunLLW18}               & 7.2M      & 610M      & 74.55          & -               & 480+50    \\
    IGCV3~\cite{SunLLW18}               & 3.5M      & 318M      & 72.2           & -               & 480+50    \\
    \specialrule{0.08em}{0pt}{0.5pt}
    \specialrule{0.08em}{0.5pt}{0pt}
    ChannelNet-v1~\cite{GaoWJ18}        & 3.7M      & 407M      & 70.5           & -               & 80    \\
    ChannelNet-v2~\cite{GaoWJ18}        & 2.7M      & -         & 69.5           & -               & 80    \\
    ChannelNet-v3~\cite{GaoWJ18}        & 1.7M      & -         & 66.7           & -               & 80    \\
    \specialrule{0.08em}{0pt}{0.5pt}
    \specialrule{0.08em}{0.5pt}{0pt}
    CondenseNet~\cite{HuangLMW18}       & 4.8M      & 529M      & 73.8           & 91.7            & 120    \\
    CondenseNet~\cite{HuangLMW18}       & 2.9M      & 274M      & 71.0           & 90.0            & 120    \\
    \specialrule{0.08em}{0pt}{0.5pt}
    \specialrule{0.08em}{0.5pt}{0pt}
    CondenseNet-FLGC~\cite{WangKSC19}   & 4.8M      & 529M      & 74.7           & 92.1            & -    \\
    \specialrule{0.08em}{0pt}{0.5pt}
    \specialrule{0.08em}{0.5pt}{0pt}
    KSE DenseNet169-A~\cite{LiLZLDWHJ19}& 7.00M     & 1.28G     & 75.79          & 92.87           & 90+21    \\
    KSE DenseNet121-A~\cite{LiLZLDWHJ19}& 4.21M     & 1.24G     & 73.9           & 91.94           & 90+21    \\
    KSE DenseNet121-B~\cite{LiLZLDWHJ19}& 3.37M     & 0.9G      & 73.03          & 91.2            & 90+21    \\
    \specialrule{0.08em}{0pt}{0.5pt}
    \specialrule{0.08em}{0.5pt}{0pt}
    NASNet-A~\cite{ZophVSL17}           & 5.3M      & 564M      & 74.0           & 91.6            & -    \\
    NASNet-B~\cite{ZophVSL17}           & 5.3M      & 488M      & 72.8           & 91.3            & -    \\
    NASNet-C~\cite{ZophVSL17}           & 4.9M      & 558M      & 72.5           & 91.0            & -    \\
    \specialrule{0.08em}{0pt}{0.5pt}
    \specialrule{0.08em}{0.5pt}{0pt}
    PNASNet-5~\cite{LiuZSHLFYHM17}      & 5.1M      & 588M      & 74.2           & 91.9            & -    \\
    \specialrule{0.08em}{0pt}{0.5pt}
    \specialrule{0.08em}{0.5pt}{0pt}
    MnasNet-A3~\cite{TanCPVL18}         & 5.2M      & 403M      & 76.7           & 93.3            & -    \\
    MnasNet-A2~\cite{TanCPVL18}         & 4.8M      & 340M      & 75.6           & 92.7            & -    \\
    MnasNet-A1~\cite{TanCPVL18}         & 3.9M      & 312M      & 75.2           & 92.5            & -    \\
    \specialrule{0.10em}{0pt}{0pt}
  \end{tabular}}
\end{table}

\section{Ablation Study}\label{sec:AblationStudy}
In this part, we conduct an ablation study to investigate the effect of the parameters such as the number of groups, the pruning step, and Conv vs. FC layers on DenseNet on the classification task of CIFAR-10/100.

\begin{figure}[!t]
  \centering
  \subfigure[]{
  \includegraphics[trim=5mm 0mm -5mm 0mm, width=2.0in]{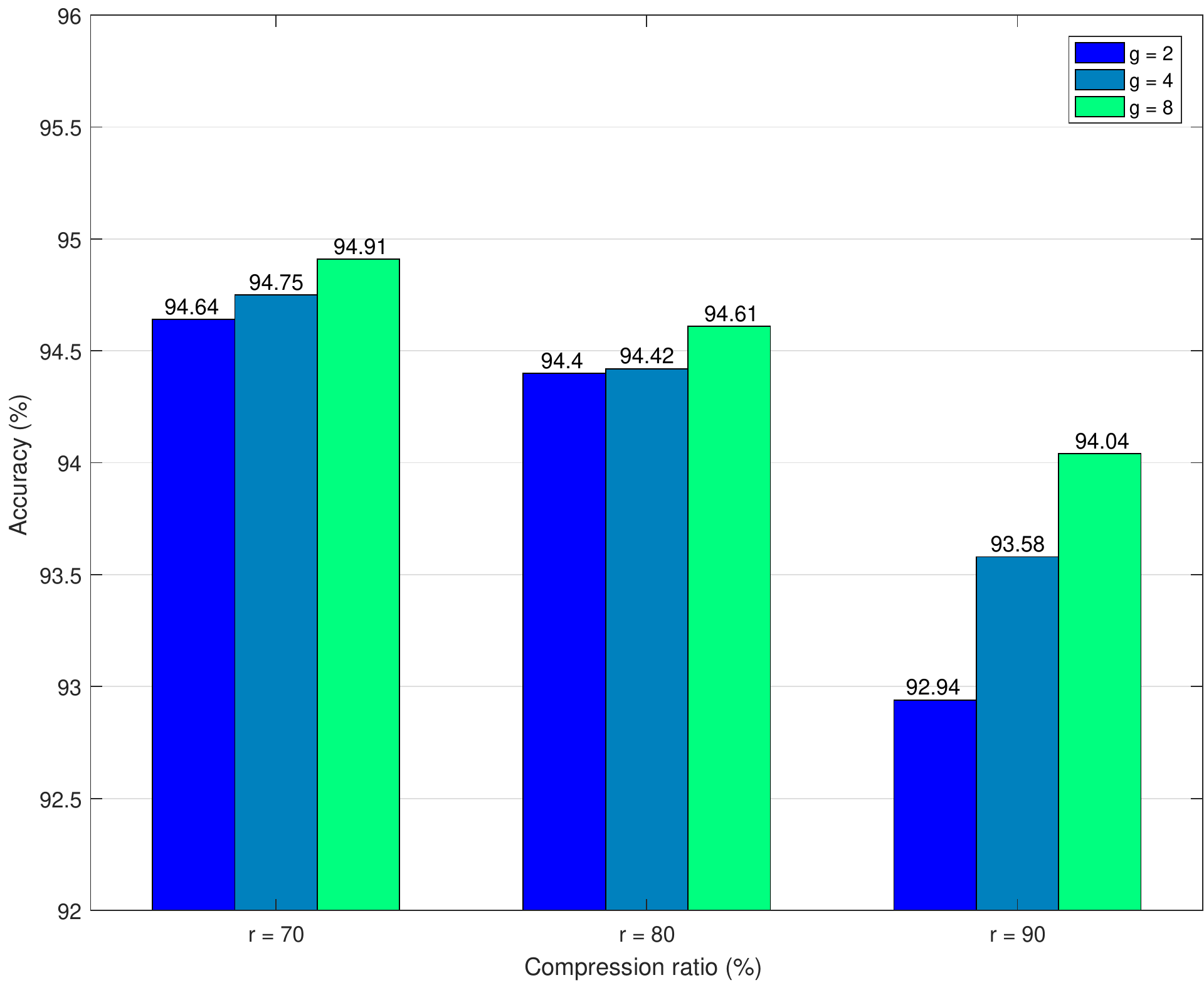}}
  \hspace{0in}
  \subfigure[]{
  \includegraphics[trim=5mm 0mm -5mm 0mm, width=2.0in]{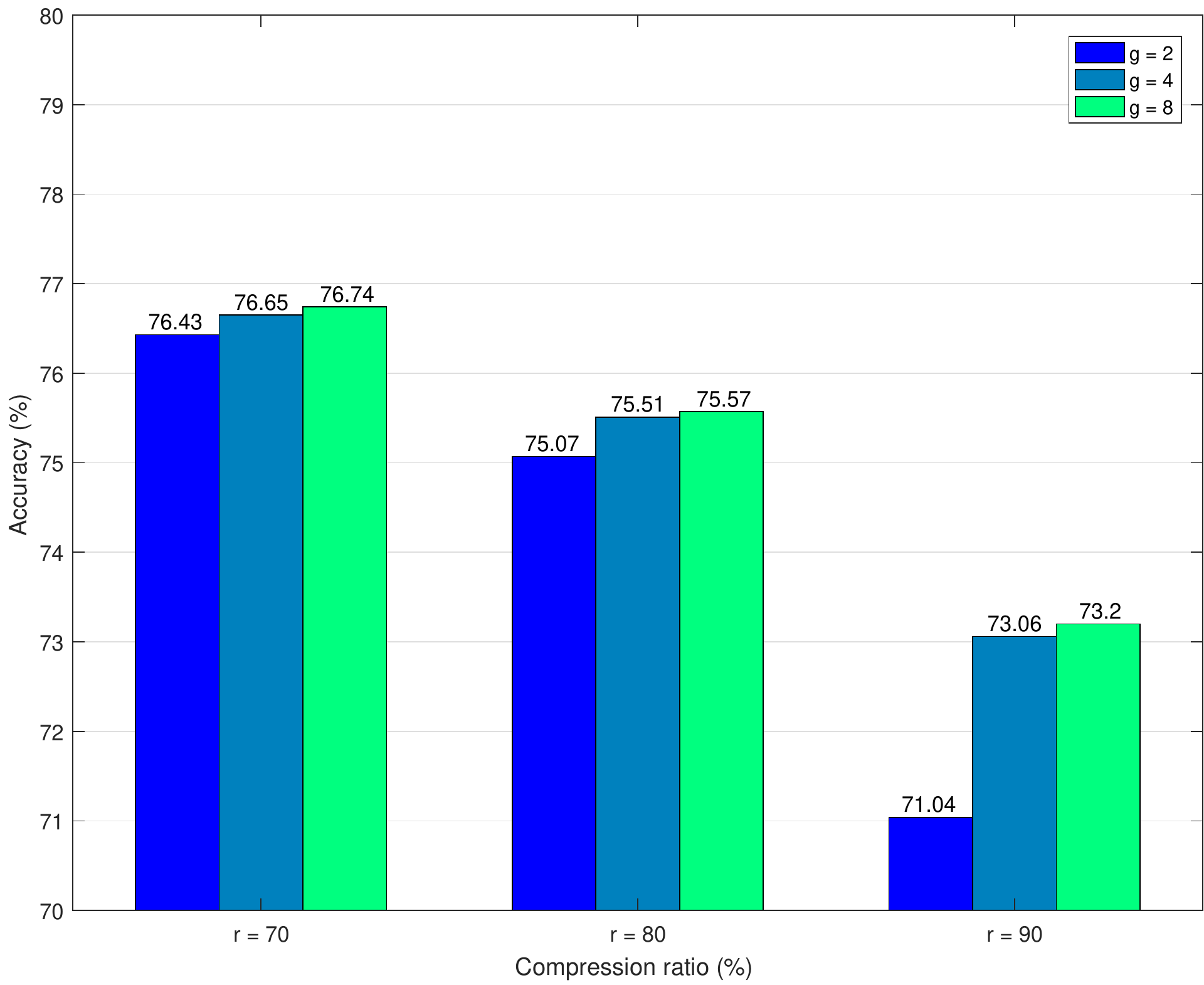}}
  {\caption{Classification accuracy (\%). (a) Accuracy vs. group number of DenseNet121 on CIFAR-10. (b) Accuracy vs. group number of DenseNet121 on CIFAR-100.}
  \label{fig:5}}
\end{figure}

\noindent
\textbf{Effect of the group number.}
Fig.~\ref{fig:5} (a) and (b) show the effect of different number of groups on DenseNet121 for CIFAR-10/100. Thanks to reusing and ignoring the shared input channels for different groups, we can have multiple group size for the same compression ratio. We fix the pruning step to 5\% for all the models, which means the same number of parameters are removed from these models each time, and further fine-tune the pruned network. From the result, we observe that a larger number of groups tends to achieve a better recognition accuracy. The gap in accuracy gradually increases with the increasing compression ratio. This suggests that increasing the number of group enhances the structural diversity of group convolutions, while preserving the information flow, which substantially improves the expressive power of the pruned networks.

\begin{figure}[!t]
  \centering
  \subfigure[]{
  \includegraphics[trim=5mm 0mm -5mm 0mm, width=2.0in]{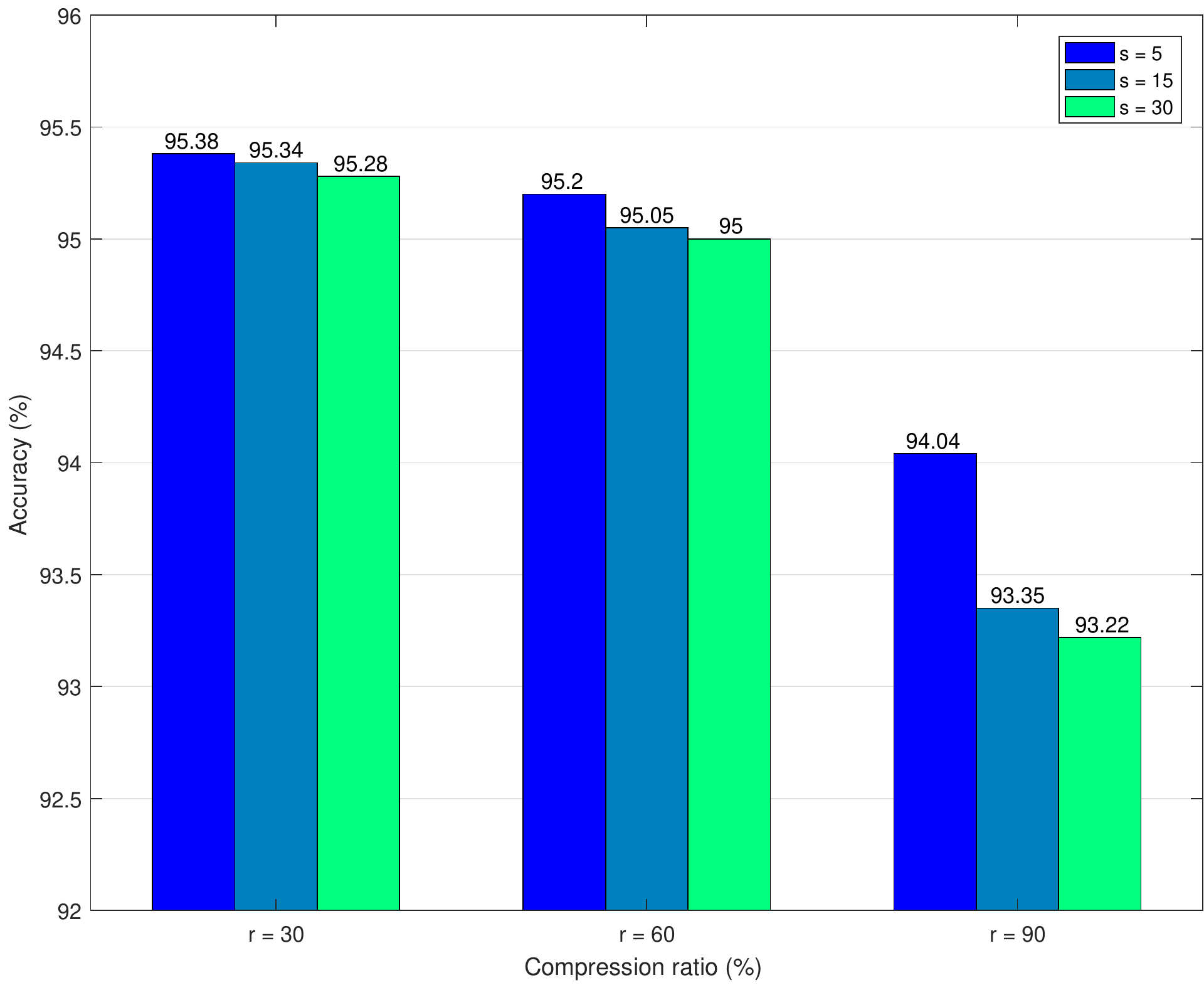}}
  \hspace{0in}
  \subfigure[]{
  \includegraphics[trim=5mm 0mm -5mm 0mm, width=2.0in]{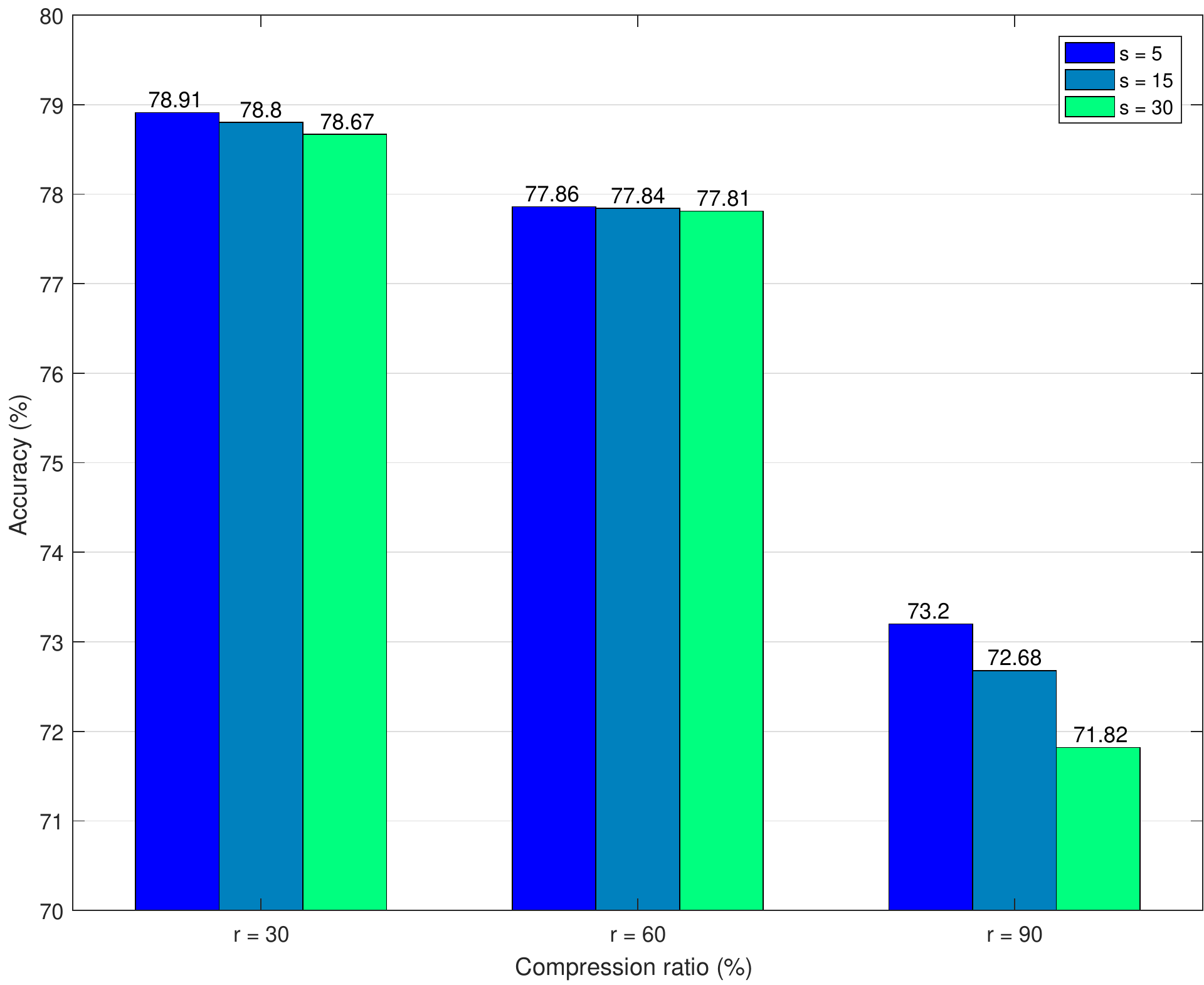}}
  {\caption{Classification accuracy (\%). (a) Accuracy vs. pruning step of DenseNet121 on CIFAR-10. (b) Accuracy vs. pruning step of DenseNet121 on CIFAR-100.}
  \label{fig:6}}
\end{figure}

\noindent
\textbf{Effect of the pruning step.}
As shown in Fig.~\ref{fig:6} (a) and (b), we illustrate the effect of different pruning steps on DenseNet121 for CIFAR-10/100. We vary the pruning step from 5\% to 30\%, and fix the number of groups to  8. The results indicate that  smaller pruning steps tend to achieve higher recognition accuracy. However, a small pruning step also affects the compression efficiency for deep neural networks. Thus, we argue that the pruning step should be set as a tradeoff between good performance and efficient compression of deep neural networks.

\begin{figure}[!t]
  \centering
  \subfigure[]{
  \includegraphics[trim=5mm 0mm -5mm 0mm, width=2.0in]{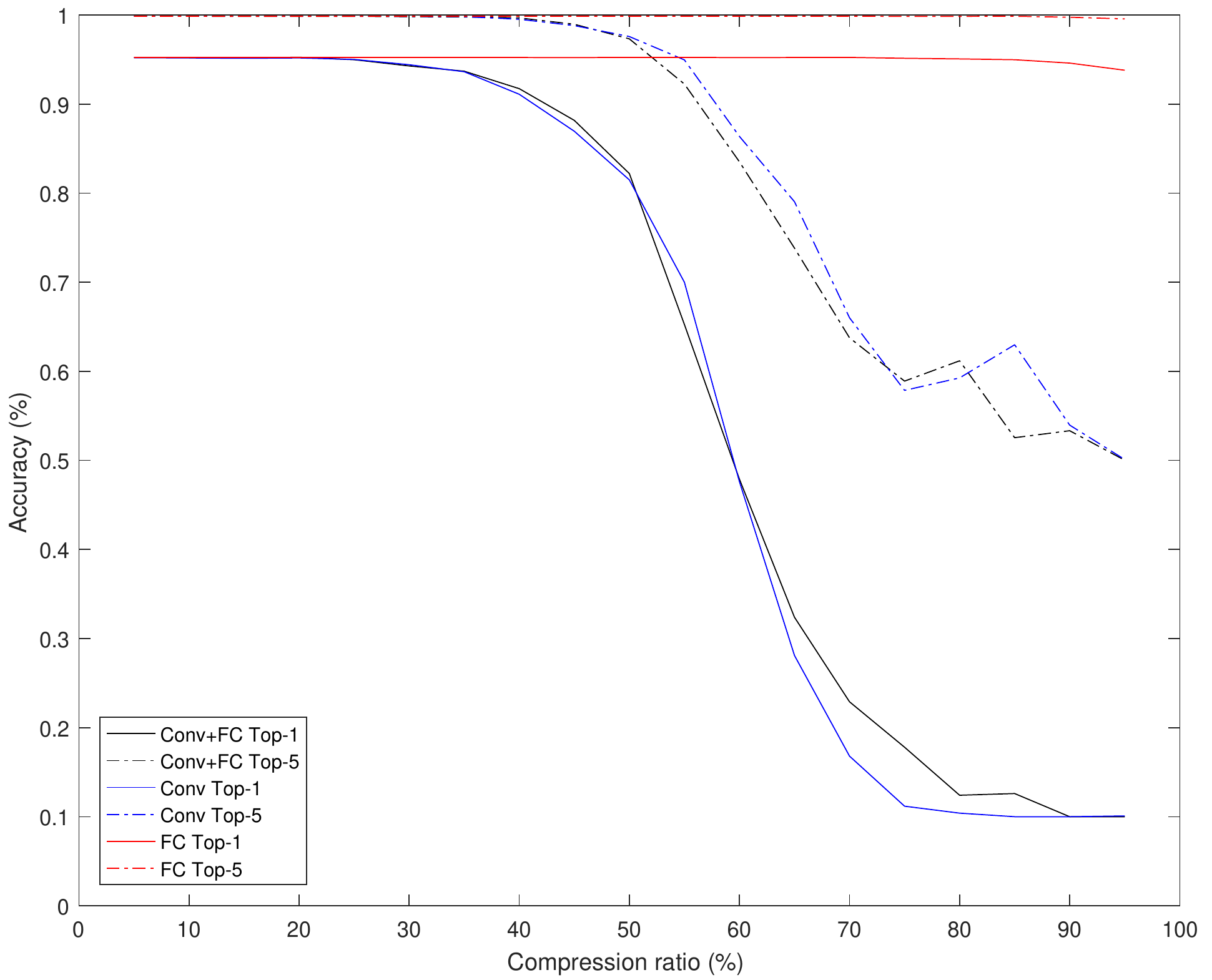}}
  \hspace{0in}
  \subfigure[]{
  \includegraphics[trim=5mm 0mm -5mm 0mm, width=2.0in]{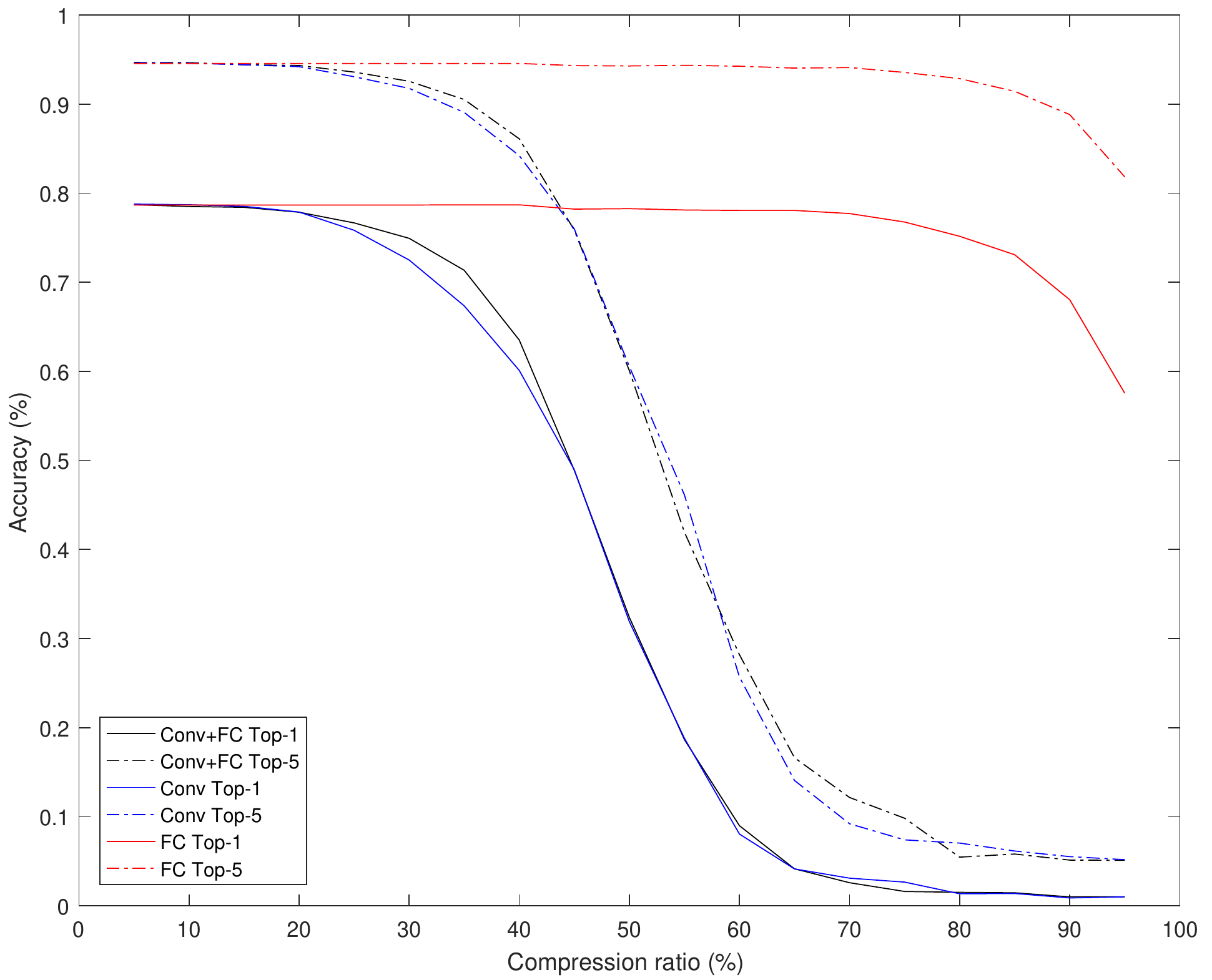}}
  {\caption{Classification accuracy (\%). (a) Accuracy vs. Conv/FC of DenseNet121 on CIFAR-10. (b) Accuracy vs. Conv/FC of DenseNet121 on CIFAR-100.}
  \label{fig:7}}
\end{figure}

\noindent
\textbf{Effect of Conv vs. FC layers.}
There are great differences between the convolutional and fully-connected layers in redundancy. To investigate their differences, we develop three different pruning schemes, i.e., pruning only Conv layers, pruning only FC layers, and pruning both of them, simultaneously. The same pruning step 5\% is set for all the models without fine-tuning. As shown in Fig.~\ref{fig:7} (a) and (b), we compare these models with different pruning schemes. All the curves remain steady when the compression ratio is less than 25\% for DenseNet121 on CIFAR-10 and 15\% for DenseNet121 on CIFAR-100, which experimentally proves that there is redundancy in these two types of layers. Afterward, the two curves of Conv and Conv+FC quickly drop with the increasing compression ratio. However, the curve of FC remains almost unchanged until its compression ratio reaches  85\% for DenseNet121 on CIFAR-10 and 65\% for DenseNet121 on CIFAR-100. So it tells us that  pruning the convolutional layer excessively can result in a degraded recognition performance. In other words, the fully-connected layers have more redundancy than the convolutional ones. Therefore, they cannot be treated  the same. Additionally, when the  compression ratio decreases, the fully-connected layer has no significant influence on the network performance. To optimise the compression ratio it is important to evaluate the degree of redundancy in the fully-connected layer.

\section{Generalization Ability}\label{sec:GeneralizationAbility}
In this section, we further evaluate the generalization ability of our SG-CNN in transfer learning, including domain adaption on CUB-200~\cite{WahCUB_200_2011} and object detection on MS COCO~\cite{LinMBHPRDZ14}. We adopt ResNet50 and DenseNet201 as our baseline models.

\noindent
\textbf{Domain Adaptation.} The CUB-200 dataset contains 11,788 images of 200 different bird species. 5,994 images are used for training and 5,794 images for testing. In order to evaluate the propensity for domain adaption, we transfer the compressed model on ImageNet into another domain, i.e., CUB-200, by fine-tuning. The same hyper-parameters and epochs are set for a fair comparison.

The result of the fine-grained classification is presented in Table~\ref{tab:6}. We observe that our SG-CNN is effective in transfer learning. The models built on ImageNet also perform well on CUB-200. In these compressed models, the models with only local fine-tuning yield the best performance, even surpassing the baseline model by a significant margin for ResNet. In comparison to both SSR~\cite{LinJLDL19}, LRDKT~\cite{LinJCTL18} and MobileNetV2~\cite{SandlerHZZC18}, our compressed models achieve even more outstanding performance for both ResNet and DenseNet. Therefore, our SG-CNN models can provide a powerful generalization to other domains or datasets.

\begin{table}[!t]
  \scriptsize
  \caption{Comparison of different compressed models for fine-grained classification on CUB-200.}
  \label{tab:6}
  \centering
  \setlength{\tabcolsep}{1.8mm}{
  \begin{tabular}{l|rr|rr}
    \specialrule{0.10em}{0pt}{0pt}
    Model &
    \makecell*[c]{Params} &
    \makecell*[c]{FLOPs} &
    \makecell*[c]{Top-1\\ (\%)} &
    \makecell*[c]{Top-5\\ (\%)} \\
    \specialrule{0.10em}{0pt}{0pt}
    Baseline                      & 23.86M  & 4.09G   & 74.37           & 94.43 \\
    ResNet-L (Conv-60/FC-60)      & 11.46M  & 1.91G   & \textbf{76.82}  & 94.96 \\
    ResNet-L (Conv-70/FC-60)      &  9.42M  & 1.56G   & 76.61           & 94.96 \\
    ResNet-L (Conv-80/FC-60)      &  7.35M  & 1.20G   & 75.18           & 94.60 \\
    \specialrule{0.08em}{0pt}{0.5pt}
    ResNet-G (Conv-60/FC-60)      & 11.47M  & 1.91G   & 73.04           & 94.01 \\
    ResNet-G (Conv-70/FC-60)      &  9.42M  & 1.55G   & 72.75           & 93.65 \\
    ResNet-G (Conv-80/FC-60)      &  7.35M  & 1.20G   & 71.92           & 92.89 \\
    \specialrule{0.08em}{0pt}{0pt}
    ResNet-LG (Conv-60/FC-60)     & 11.46M  & 1.91G   & 73.25           & 93.63 \\
    ResNet-LG (Conv-70/FC-60)     &  9.42M  & 1.56G   & 73.11           & 93.22 \\
    ResNet-LG (Conv-80/FC-60)     &  7.35M  & 1.20G   & 71.94           & 93.32 \\
    \specialrule{0.08em}{0.5pt}{0pt}
    \specialrule{0.08em}{0pt}{0pt}
    Baseline                      & 18.28M  & 4.29G   & 78.65           & 95.46 \\
    DenseNet-L (Conv-70/FC-60)    &  5.61M  & 1.34G   & \textbf{77.93}  & \textbf{95.44} \\
    DenseNet-L (Conv-80/FC-60)    &  3.93M  & 0.94G   & 77.17           & 95.05 \\
    \specialrule{0.08em}{0pt}{0pt}
    DenseNet-G (Conv-70/FC-60)    &  5.66M  & 1.35G   & 77.20           & 94.70 \\
    DenseNet-G (Conv-80/FC-60)    &  3.94M  & 0.94G   & 75.73           & 94.25 \\
    \specialrule{0.08em}{0pt}{0pt}
    DenseNet-LG (Conv-70/FC-60)   &  5.61M  & 1.34G   & 77.46           & 94.98 \\
    DenseNet-LG (Conv-80/FC-60)   &  3.93M  & 0.94G   & 75.66           & 94.56 \\
    \specialrule{0.08em}{0pt}{0.5pt}
    \specialrule{0.08em}{0.5pt}{0pt}
    SSR-L2,1~\cite{LinJLDL19}     & 124.6M  & 4.5G    & 71.30           & - \\
    SSR-L2,1-GAP~\cite{LinJLDL19} &   8.8M  & 4.4G    & 70.45           & - \\
    \specialrule{0.08em}{0pt}{0pt}
    LRDKT~\cite{LinJCTL18}        &   9.5M  & 1.31G   & 75.10           & - \\
    LRDKT~\cite{LinJCTL18}        &   3.7M  & 0.64G   & 63.18           & - \\
    \specialrule{0.08em}{0pt}{0pt}
    MobileNetV2~\cite{SandlerHZZC18}~(Our impl.)   & 2.45M & 0.30G& 68.85         & 91.75 \\
    \specialrule{0.10em}{0pt}{0pt}
  \end{tabular}}
\end{table}

\noindent
\textbf{Object Detection.} To evaluate the ability to detect objects, we deploy the compressed model over Faster R-CNN~\cite{RenHGS15} as the detection framework, and use the publicly released pytorch code~\cite{jjfaster2rcnn} for implementation with default settings. The models are trained on COCO train+val dataset excluding 5K minival images, and evaluated on the minival set, with 300 and 600 input resolutions.

Table~\ref{tab:7} shows the comparison results on the two input resolutions. For both network frameworks, compared to their baseline models, our SG-CNN offers $>$3$\times$ smaller model size and $\approx$3.5$\times$ lower FLOPs for ResNet50 and $>$4$\times$ smaller model size and $>$4.5$\times$ lower FLOPs for DenseNet201, while obtaining good or even better performance in  object detection. Our models with only global fine-tuning achieve comparable object detection results to the competitors with both local and global fine-tuning, showing that our method preserves the flow of relevant information at each layer after pruning. Both MobileNet~\cite{HowardZCKWWAA17} and ShuffleNet~\cite{ZhangZLS17} are outperformed by  our self-grouping method by a significant margin on both resolutions. It is also apparent that our  method exhibits excellent generalization ability in object detection.

\begin{table}[!t]
  \scriptsize
  \caption{The object detection results on MS COCO. Here, mAP-1 and mAP-2 correspond to 300$\times$ and 600$\times$ input resolutions, respectively. mAP is reported with COCO primary challenge metric (AP@IoU=0.50:0.05:0.95).}
  \label{tab:7}
  \centering
  \setlength{\tabcolsep}{1.8mm}{
  \begin{tabular}{l|rr|rr}
    \specialrule{0.10em}{0pt}{0pt}
    Model &
    \makecell*[c]{Params} &
    \makecell*[c]{FLOPs} &
    \makecell*[c]{mAP-1\\ (\%)} &
    \makecell*[c]{mAP-2\\ (\%)} \\
    \specialrule{0.10em}{0pt}{0pt}
    Baseline                          & 24.44M & 4.09G   & 24.5          & 30.9 \\
    \specialrule{0.08em}{0pt}{0pt}
    ResNet-G (Conv-60/FC-60)          & 12.00M & 1.91G   & 24.8          & \textbf{30.9} \\
    ResNet-G (Conv-70/FC-60)          &  9.95M & 1.55G   & 24.2          & 29.5 \\
    ResNet-G (Conv-80/FC-60)          &  7.88M & 1.20G   & 23.2          & 28.6 \\
    \specialrule{0.08em}{0pt}{0pt}
    ResNet-LG (Conv-60/FC-60)         & 11.99M & 1.91G   & \textbf{24.9} & \textbf{30.9} \\
    ResNet-LG (Conv-70/FC-60)         &  9.95M & 1.56G   & 24.2          & 30.0 \\
    ResNet-LG (Conv-80/FC-60)         &  7.88M & 1.20G   & 23.0          & 28.2 \\
    \specialrule{0.08em}{0pt}{0.5pt}
    \specialrule{0.08em}{0.5pt}{0pt}
    Baseline                          & 18.78M & 4.29G   & 26.0          & 32.8 \\
    \specialrule{0.08em}{0pt}{0pt}
    DenseNet-G (Conv-70/FC-60)        &  6.15M & 1.35G   & 23.9          & 30.3 \\
    DenseNet-G (Conv-80/FC-60)        &  4.44M & 0.94G   & 22.7          & 28.7 \\
    \specialrule{0.08em}{0pt}{0pt}
    DenseNet-LG (Conv-70/FC-60)       &  6.11M & 1.34G   & \textbf{24.0} & \textbf{30.3} \\
    DenseNet-LG (Conv-80/FC-60)       &  4.43M & 0.94G   & 23.0          & 28.9 \\
    \specialrule{0.08em}{0pt}{0.5pt}
    \specialrule{0.08em}{0.5pt}{0pt}
    MobileNetV1~\cite{HowardZCKWWAA17}& 4.25M  & 516.80M & 16.4          & 19.8 \\
    \specialrule{0.08em}{0pt}{0pt}
    ShuffleNetV1~\cite{ZhangZLS17}    & 4.25M  & 516.80M & 18.7          & 25.0 \\
    \specialrule{0.10em}{0pt}{0pt}
  \end{tabular}}
\end{table}


\section{Conclusion}\label{sec:Conclusion}
We have presented a self-grouping convolutional neural network, named SG-CNN, to improve the existing group convolution methods for the compression and acceleration of deep neural networks, for the deployment on mobile and embedded devices with constrained memory and computation. We automatically group the filters for each convolutional layer by clustering based on the importance vectors, and further enhance sparsity of each group by pruning based on their cluster centroids, thus yielding a self-grouping convolution which is data-dependent and has diverse group structures. Furthermore, our SG-CNN works throughout the fully-connected layers as well as the convolutional layers, aiming to simultaneously accelerate inference and reduce memory consumption. Both local fine-tuning and global tuning further improve the recognition accuracy of the pruned network. We empirically demonstrated the effectiveness and efficiency of our approach on a variety of state-of-the-art CNN architectures, such as ResNet and DenseNet, on four popular datasets, including CIFAR-10/100 and ImageNet. The experimental results show our self-grouping method achieves superior performance. Particularly, for ResNet50, our SG-CNN achieves over 3$\times$ compression rate and about 3.5$\times$ FLOPs reduction with 73.38\% top-1 accuracy and 91.69\% to-5 accuracy on ImageNet. For DenseNet201, our SG-CNN achieves over 4.5$\times$ compression rate and over 4$\times$ FLOPs reduction, while delivering  75.17\% top-1 recognition accuracy and 92.6\% top-5 accuracy on ImageNet. We further evaluated the generalization ability of SG-CNN on both domain adaption and object detection, and obtained competitive results.

\section*{Acknowledgment}
This work is supported by the National Key R\&D Program of China (Grant No. 2018YFB1004901), and the Independent Innovation Team Project of Jinan City (No. 2019GXRC013), by the National Natural Science Foundation of China (Grant No.61672265, U1836218), by the 111 Project of Ministry of Education of China (Grant No. B12018), by UK EPSRC GRANT EP/N007743/1, MURI/EPSRC/DSTL GRANT EP/R018456/1.


\bibliographystyle{elsarticle-num}
\bibliography{elsarticle-template-num}





%
%
%
\end{document}